%% file: article.tex
\documentclass[review,onefignum,onetabnum]{siamonline250106}

\input{shared}

\ifpdf
\hypersetup{
  pdftitle={G-Mapper: Learning a Cover in the Mapper Construction},
  pdfauthor={E. Alvarado, R. Belton, E. Fischer, K. Lee, S. Palande, S. Percival, and E. Purvine}
}
\fi

\begin{document}
\nolinenumbers
\maketitle

\input{body/abstract.tex}



\input{body/intro.tex}
\input{body/background.tex}

\input{body/method.tex}
\input{body/results.tex}
\input{body/discussion.tex}

\section*{Acknowledgments}
The authors would like to thank the anonymous reviewers for their careful reading of the manuscript and their many insightful comments and suggestions. The authors would like to thank Nithin Chalapathi, Youjia Zhou, and Bei Wang for their encouraging conversations and their comments concerning how to handle datasets. The authors would also like to thank Sarah Tymochko for helpful input during early conversations of this work.
\bibliographystyle{siamplain}
\bibliography{references}

\end{document}

%% file: shared.tex
\usepackage[utf8]{inputenc}
\usepackage[margin=1in]{geometry}
\usepackage{appendix}
\usepackage{graphicx}
\usepackage{wrapfig}
\usepackage{multicol}
\usepackage{mathrsfs}  
\usepackage{vwcol}
\usepackage{amsbsy}
\usepackage[normalem]{ulem}
\usepackage{tikz-cd}
\usepackage{tikz}
\usepackage{extarrows}
\usepackage{amsmath,amssymb}
\usepackage{hyperref}
\usepackage{tcolorbox}
\usepackage{color}
\usepackage{subcaption}

\definecolor{darkorchid}{rgb}{0.6,0.196,0.8}
\definecolor{sky}{RGB}{8,133,196}

\hypersetup{
colorlinks = true,
linkcolor = violet,
citecolor = violet,
urlcolor = violet
}

\newcommand{\R}{\mathbb{R}}

\newcommand{\goverlap}{{\fontfamily{cmtt}\selectfont g\_overlap }}

\newcommand{\figref}[1]{Figure~\ref{fig:#1}}
\newcommand{\tableref}[1]{Table~\ref{table:#1}}

\newcommand{\exref}[1]{Example~\ref{ex:#1}}

\newcommand{\secref}[1]{Section~\ref{sec:#1}}

\ifpdf
  \DeclareGraphicsExtensions{.eps,.pdf,.png,.jpg}
\else
  \DeclareGraphicsExtensions{.eps}
\fi

\newtheorem{thm}{Theorem}[section]
\newtheorem{example}[thm]{Example}{\bfseries}{\itshape}

\headers{G-Mapper: Learning a Cover in the Mapper Construction}{E. Alvarado, R. Belton, E. Fischer, K. Lee, S. Palande, S. Percival, and E. Purvine}

\title{G-Mapper: Learning a Cover in the Mapper Construction\thanks{{\bfseries Funding:} This research is a product of one of the working groups at the American Mathematical Society (AMS) Mathematical Research Community: \textit{Models and Methods for Sparse (Hyper)Network Science} in June 2022. The workshop and follow-up collaboration was supported by the National Science Foundation under grant number DMS 1916439. Any opinions, findings, and conclusions or recommendations expressed in this
material are those of the author(s) and do not necessarily reflect the views of the National Science Foundation or the American Mathematical Society. Kang-Ju Lee was supported in part by the National Research Foundation of Korea (NRF) Grants funded by the Korean
Government (MSIP) (No.2021R1C1C2014185).}}

\author{Enrique Alvarado\thanks{Department of Mathematics, Iowa State University, Ames, IA, USA 
  (\email{enrique3@iastate.edu}).}
\and 
Robin Belton\thanks{Department of Mathematics and Statistics, Vassar College, Poughkeepsie, NY, USA (\email{rbelton@vassar.edu}).}
\and 
 Emily Fischer\thanks{Umqua Bank, Portland, OR, USA (\email{EmilyFischer@UmpquaBank.com}).}
\and 
Kang-Ju Lee\thanks{Research Institute of Mathematics, Seoul National University, Seoul, KR (\email{leekj0706@snu.ac.kr}).}
\and 
 Sourabh Palande
 \thanks{Donald Danforth Plant Science Center, Olivette, MO, USA (\email{spalande@danforthcenter.org}).} 
\and 
Sarah Percival\thanks{Department of Mathematics and Statistics, University of New Mexico, Albuquerque, NM, USA (\email{spercival@unm.edu}).}
\and 
Emilie Purvine\thanks{Mathematics of Data Science, Pacific Northwest National Laboratory, Richland, WA, USA (\email{Emilie.Purvine@pnnl.gov)}.}}
\usepackage{amsopn}

%% file: body/abstract.tex
\begin{abstract}
The Mapper algorithm is a visualization technique in topological data analysis (TDA) that outputs a graph reflecting the structure of a given dataset. However, the Mapper algorithm requires tuning several parameters in order to generate a ``nice" Mapper graph. This paper focuses on selecting the cover parameter. We present an algorithm that optimizes the cover of a Mapper graph by splitting a cover repeatedly according to a statistical test for normality. Our algorithm is based on G-means clustering which searches for the optimal number of clusters in $k$-means by iteratively applying the Anderson-Darling test. Our splitting procedure employs a Gaussian mixture model to carefully choose the cover according to the distribution of the given data. Experiments for synthetic and real-world datasets demonstrate that our algorithm generates covers so that the Mapper graphs retain the essence of the datasets, while also running significantly faster than a previous iterative method.  
\end{abstract}

%% file: body/intro.tex
\section{Introduction} \label{sec:intro}

Topological data analysis (TDA) utilizes techniques from topology in order to extract valuable insights from a dataset. Topology, the major branches of mathematics, studies the shape of a space, and TDA uncovers the shape of the dataset using topology. We refer the reader to \cite{carlsson2009topology} for an overview of this area. This paper focuses on the \emph{Mapper} algorithm introduced in \cite{singh2007topological} by Singh, Mémoli, and Carlsson, one of the fundamental visualization tools in TDA. Mapper has been shown to be useful in various applications such as analyzing breast cancer microarray data \cite{nicolau2011topology}, identifying diabetes subgroups \cite{li2015identification}, and studying divergence of COVID-19 trends \cite{zhou2021mapper}. 

Mapper is a network-based visualization technique of high-dimensional data. The algorithm takes as input a point cloud dataset and produces as output a graph (or network) reflecting the structure of the underlying data. To apply the Mapper algorithm, the user needs to determine the following parameters that include choosing a \emph{lens} (or \emph{filter}) function $f:X \to Y$ from a high-dimensional point cloud $X$ to a lower-dimensional space $Y$, an \emph{(open) cover} of the target space $Y$, and a \emph{clustering} algorithm for cover elements. Optimizing these parameters is essential for generating a ``nice'' Mapper graph, i.e. a graph that captures the underlying structure of the dataset and, in the case where the dataset is well-studied, is consistent with the literature. We concentrate on tuning a \emph{cover} given by a collection of overlapping intervals. While the traditional Mapper algorithm takes a uniform cover with the number of intervals and the same overlapping percent between consecutive intervals to be specified by the user \cite{singh2007topological}, sophisticated methods have recently been applied to optimize the cover of Mapper.
\vskip 10pt

This paper is concerned with \emph{clustering} methods which have been utilized for selection of a cover. Motivated by the \emph{$X$-means} algorithm \cite{pelleg2000x} for estimating the number of clusters in \emph{k-means} according to the \emph{Bayesian Information Criterion} (BIC), Chalapathi, Zhou, and Wang devised an algorithm which repeatedly splits intervals of a coarse cover \cite{chalapathi2021adaptive} according to information criteria. This Mapper algorithm is called \emph{Multipass AIC/BIC}. Our work is primarily inspired by this algorithm.  Additionally, employing the \emph{fuzzy $c$-means} algorithm \cite{dunn1973fuzzy, bezdek2013pattern}, a centroid-based overlapping clustering method, Bui et al. presented a Mapper construction called \emph{F-Mapper} that takes clusters obtained by the algorithm applied to $f(X)$ as a cover of Mapper \cite{bui2020f}. We develop a Mapper algorithm based on another iterative clustering algorithm (\emph{G-means} \cite{hamerly2003learning}) utilizing a statistical test.

\vskip 15pt
\subsection{Contributions}

We propose a new Mapper construction algorithm called \emph{G-Mapper} for optimizing a cover of the Mapper graph based on G-means clustering \cite{hamerly2003learning}. The G-means clustering algorithm aims to learn the number $k$ of clusters in the $k$-means clustering algorithm according to a statistical test, called the \emph{Anderson–Darling} test, for the hypothesis that the points in a cluster follow a \emph{Gaussian} distribution. Our algorithm splits a cover element iteratively, with our splitting decision determined by the Anderson–Darling score. For further elaboration, we split each cover element into two overlapping intervals employing a \emph{Gaussian mixture model} (GMM) so that the splits are made according to the characteristics of the cover element rather than producing uniform intervals. The GMM is a soft clustering technique that determines the probability of a given data point belonging to a cluster, making it well-suited for constructing overlapping intervals. The procedure allows us to take variance into account when forming cover elements, making our algorithm perform well without the initialization of a cover.

\vskip 10pt

G-Mapper integrates ideas from the current state-of-the-art techniques to fill in gaps from each method. To find the number of intervals in a cover, the Multipass AIC/BIC method iteratively splits cover elements using information criteria. However, experiments reveal that $X$-means based on the criteria does not perform well when a given dataset is non-spherical or high-dimensional \cite{hamerly2003learning,hu2004investigation,sinaga2020unsupervised} and it has computational issues
\cite{ikotun2023k}. Additionally, when a cover element splits, two intervals are created to have a set uniform length. In contrast, F-Mapper takes into account the data to determine where intervals in the cover should be located, but the number of intervals must be predetermined. G-Mapper finds the number of intervals by applying an iterative splitting procedure with the Anderson-Darling test. G-means works well experimentally even for non-spherical data and high-dimensional data, performing considerably faster than $X$-means \cite{hamerly2003learning}. Moreover, our method splits a cover element into two overlapping intervals using a GMM, considering the distribution of the data.
\vskip 10pt
Applying the G-Mapper algorithm to various synthetic and real-world datasets, we demonstrate that the algorithm generates covers such that the corresponding Mapper graphs preserve the important structural features of the datasets, as determined by previous studies of the same datasets. A comparison of Mapper graphs, generated by G-Mapper and Multipass BIC, indicates that our algorithm captures characteristics of the datasets that are not detected by the other algorithms, performs better even for high-dimensional datasets, and runs significantly faster. Experiments also reveal that while Multipass BIC requires initialization of a cover, G-Mapper does not due to utilizing a GMM. In addition, we illustrate that the number of intervals in the cover produced by G-Mapper could be utilized as input for other Mapper algorithms such as F-Mapper. We provide an open-source implementation (\url{https://github.com/MRC-Mapper/G-Mapper}) and we are working on distributing it as a stand-alone Python package.

\subsection{Related Work}

Following the original Mapper construction algorithm \cite{singh2007topological}, most implementations of Mapper including \cite{van2019kepler, tauzin2020giotto} construct the cover using open intervals or hypercubes of a fixed length with a fixed amount of overlap. However, the optimal number of intervals or amount of overlap to use is often unknown and the Mapper output is very sensitive to these parameters. 

The implementation of Mapper in Giotto TDA \cite{tauzin2020giotto} has a \emph{balanced cover} method in which the user specifies the number of intervals and the algorithm finds a cover with the given number of intervals so that each interval covers the same number of points. This parameter is also difficult to know a priori and the method may not be elaborate on generating an optimal cover.

Statistical methods are utilized for optimizing a cover of the Mapper construction algorithm. Carri\`{e}re, Michel, and Oudot introduced a statistical method to select these parameters \cite{carriere2018statistical}. The main idea is to sweep through various Mapper parameters to find the ``best” Mapper graph that is structurally stable and close to a particular \emph{Reeb graph} \cite{reeb46points}. \emph{Extended persistence} \cite{cohen2009extending} developed by Cohen-Steiner, Edelsbrunner, and Harer is used to find this optimal Mapper graph. This method requires the user to know the Reeb graph to compare the Mapper graph against and relies on independent sampling conditions of the point cloud. Vejdemo-Johansson and Leshchenko \cite{vejdemo2020certified} developed \emph{Certified Mapper}, a statistical method for producing certificates that identifies the non-obstruction to the Nerve Lemma.

Clustering methods such as fuzzy $c$-means \cite{dunn1973fuzzy, bezdek2013pattern} and $X$-means \cite{pelleg2000x} have been used for optimizing an open cover \cite{bui2020f, chalapathi2021adaptive}. These methods were already mentioned in the third paragraph of \secref{intro}. The cover optimizing algorithm we propose is motivated by a clustering method called the G-means clustering \cite{hamerly2003learning}.

%% file: body/background.tex
\section{Background}

This section is devoted to reviewing the Mapper construction and statistical tools used in our proposed G-Mapper algorithm to tune the cover parameter. \secref{mapper} describes the Mapper construction and we focus on a cover constructed from a set of points in $\R$. The central idea of our algorithm is to form cover elements (intervals) around points that appear to be normally distributed. In order to test if a set of points in an interval follows a Gaussian distribution, we apply the Anderson-Darling test that we explain in \secref{anderson-darling}. If the statistical test indicates the points do not follow a Gaussian distribution, we split the interval into two intervals using a Gaussian mixture model reviewed in \secref{gmm}. In \secref{g-means}, we describe the G-means clustering algorithm that inspires our G-Mapper algorithm.
\vskip 10pt
\subsection{Mapper}\label{sec:mapper}
The Mapper algorithm was first introduced in \cite{singh2007topological} and generates a graph or a network. The output is called a Mapper graph. For a given dataset $X$, constructing a Mapper graph consists of the following procedure that we illustrate in \exref{mapper}.

\begin{enumerate}
\item Define a lens (or filter) function $f: X \to Y$ from the point cloud $X$ to a lower-dimensional space $Y$. In this paper, we focus on the case that the target space $Y$ is a subset of $\R$, i.e., the space $Y$ is of dimension $1$.

\item Construct a cover $\mathcal{U}=\{\,U_i\,|\, U_i \mbox{ is open for }i \in I\}$ of the target space $Y$ given in Step 1, i.e., \mbox{$Y \subset \bigcup_{i \in I} U_i$}. For a one-dimensional space $Y$, a cover of $Y$ consists of overlapping intervals. 

\item For each element $U_i$ of the cover, apply a clustering algorithm to the pre-image $f^{-1}(U_i)$ of $U_i$ under the lens function $f$.

\item Create the Mapper graph whose vertices are the clusters found in Step 3, and an edge between two vertices exists if the two corresponding clusters share data points. The output is a simplified representation of the dataset $X$. Note that the Mapper graph corresponds to the $1$-skeleton of the nerve of the cover of $X$ generated in Step 3.

\end{enumerate}
\vskip 20pt
In Step 2, the conventional Mapper algorithm \cite{singh2007topological} employs covers consisting of intervals of uniform length and the overlap G between each two consecutive intervals. The parameter G and the number $r$ of intervals are referred to as the \emph{gain} and the \emph{resolution}, respectively. The user should tune these two parameters to optimize a cover in the conventional Mapper algorithm.

In Step 3, we use a density-based clustering algorithm called DBSCAN \cite{ester1996density}. The algorithm is popularly chosen for Mapper because it does not require the number of clusters to be pre-determined and it detects arbitrarily shaped clusters. We remove outliers identified by the algorithm since they do not significantly contribute to the overall structure of the dataset, which is what we aim to recover in the Mapper graph. Recall that DBSCAN has two main parameters: $\varepsilon$, the maximum distance at which a sample can still be considered part of another's neighborhood, and \text{{\fontfamily{cmtt}\selectfont MinPts}}, the minimum number of samples needed in a neighborhood for a point to be considered as a core point.

\begin{figure}[htp]
    \centering
    \begin{subfigure}[b]{0.235\textwidth}
        \centering
        \includegraphics[width=.85\textwidth]{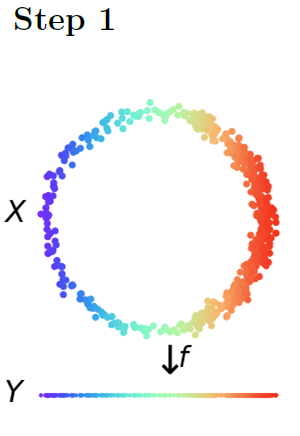}
        \caption{Lens Function}
        \label{fig:filter}
    \end{subfigure}
    \hfil
    \begin{subfigure}[b]{0.235\textwidth}
        \centering
        \includegraphics[width=.85\textwidth]{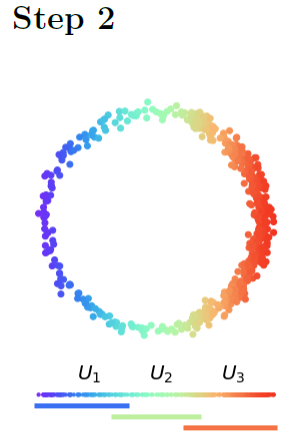}
        \caption{Cover}
        \label{fig:cover}
    \end{subfigure}
    \hfil
    \begin{subfigure}[b]{0.235\textwidth}
        \centering
        \includegraphics[width=.85\textwidth]{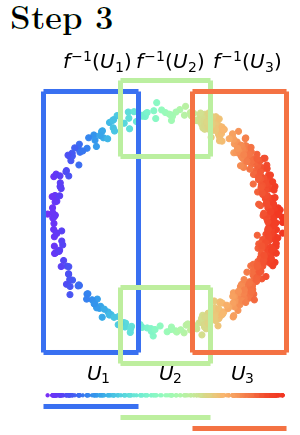}
        \caption{Clustering Pre-images}
        \label{fig:clustering}
    \end{subfigure}
        \begin{subfigure}[b]{0.235\textwidth}
        \centering
        \includegraphics[width=.85\textwidth]{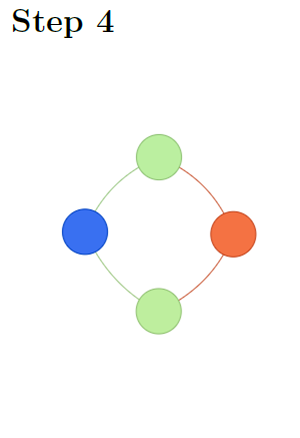}
        \caption{Mapper Graph}
        \label{fig:Mapper}
    \end{subfigure}
    \caption{Mapper Construction. The dataset $X$ consists of points sampled from a circle of radius $1/2$ with center at $(1/2,1/2)$. Constructing a Mapper graph requires selecting a lens function $f$ (\figref{filter}) and cover $\{U_i\}$ (\figref{cover}), and applying a clustering algorithm to $\{f^{-1}(U_i)\}$ (\figref{clustering}). The parameters are specified in \exref{mapper}, and the generated Mapper graph is a cycle graph with four vertices and four edges (\figref{Mapper}).}
    \label{fig:Mapper_Construction}
\end{figure}

We illustrate the Mapper construction algorithm by applying it to a toy dataset (circle dataset) in \exref{mapper}. 
\begin{example}[Mapper Graph]\label{ex:mapper}
\normalfont Let $X$ be the set of points sampled from a circle of radius $1/2$ with center at $(1/2,1/2)$. Refer to \figref{Mapper_Construction}.
\begin{enumerate}
\item We define the lens function $f: X\to Y$ to be the projection map onto the first coordinate with the target space $Y = [-0.03, 1.05]$. 
\item The cover $\{U_i\}_{i=1}^3$ is chosen to be three overlapping intervals of $Y$. Specifically, we set $U_1=[-0.03,0.39)$, $U_2=(0.30,0.72)$, and $U_3=(0.64,1.05]$. Consecutive intervals are overlapping by $20\%$.
\item The DBSCAN clustering algorithm generates four clusters where each of $f^{-1}(U_1)$ and $f^{-1}(U_3)$ forms a cluster and $f^{-1}(U_2)$ is separated into two clusters based on whether the points are in the upper or lower part of the circle. 
\item The Mapper graph constructed in Step 4 is a cycle graph with four vertices and four edges. The color of each vertex in the Mapper graph is the average value of the lens function with the rainbow color map. 
\end{enumerate}
\end{example}

\subsection{Statistical Test for a Gaussian Distribution} \label{sec:anderson-darling}
The Anderson-Darling test is a statistical test to determine if data follows a Gaussian distribution \cite{anderson1952asymptotic, stephens1974edf}. The test is similar to two other normality tests, the Kolmogorov-Smirnov test \cite{massey1951kolmogorov} and the Shapiro-Wilk test \cite{shapiro1965analysis} in that it is computing test statistics based on the empirical distribution function (EDF).

Let $X=\{x_i\}_{i=1}^n$ be a given dataset. We standardize the data to yield a set $X'=\{x'_i\}_{i=1}^n$ such that $X'$ has a mean of $0$ and a variance of $1$. Let $x'_{(i)}$ be the $i^{th}$ ordered value of $X'$. Define $z_i=F(x'_{(i)})$, where $F$ is the cumulative distribution function of the standard normal distribution. Let $F_n$ be the sample cumulative distribution function. Then the \emph{Anderson-Darling (AD)} statistic is defined to be the quadratic EDF statistic measuring differences between $F$ and $F_n$:
\[
A^2(X) = n\int_{-\infty}^{\infty}\left(F_n(x)-F(x)\right)^2w(x)\mbox{d}F(x),
\] \vskip 1pt
where $w(x)$ is a weighting function given by $w(x)=(F(x)(1-F(x)))^{-1}$. This weighting function places more weight on observations in the tails. Computing the integration results in 
\[
A^2(X) = -\frac{1}{n}\sum_{i=1}^n(2i-1)\left(\log\left(z_i\right)+\log\left(1-z_{n+1-i}\right)\right)-n.
\] \vskip 1pt
In the case where both the mean and the variance are unknown and estimated from the data, the following modification to the statistic is known to provide more accurate results \cite{stephens1974edf}:
\[
A_{*}^2(X) = A^2(X)(1+4/n - 25/n^2).
\] \vskip 1pt
The user selects a critical threshold, which we call the \textit{AD threshold}. The set of AD statistics exceeding the critical threshold is called the critical region. The null hypothesis $H_0$, asserting that the data follows a Gaussian distribution, is rejected for values $A_*^2(X)$ in the critical region, and $H_0$ is not rejected for values below the AD threshold. The probability of $A_*^2$ occurring in the critical region under the null hypothesis is the \textit{significance level} $\alpha$. A table of critical values for varying significance levels can be found in \cite{d2017tests}.
\vskip 15pt
\subsection{Gaussian Mixture Models} \label{sec:gmm}
A Gaussian mixture model (GMM) is a probabilistic model representing a probability density function as a finite weighted sum of Gaussian distributions. It is known to be effective for modeling complex data distributions by combining multiple Gaussian components. The model learns a mixture model distribution with $k$ components. The probability density function $p(x)$ of a real-valued vector $x$ is given by
\[
p(x) = \sum_{i=1}^k \pi_i \,N[ x | m_i, \Sigma_i],
\] \vskip 1pt
where $\pi_i$ is a mixture probability with $\sum_{i=1}^{k}\pi_i =1$, and $N[x | m_i, \Sigma_i]$ is the $i^{th}$ Gaussian distribution with mean $m_i$ and covariance $\Sigma_i$. For further details on this model, see Section 2.3.9 and Section 9.2.2 of \cite{bishop2006pattern} and refer to \exref{AD_GMM} for an illustration.

To fit the model, the GMM implements the expectation-maximization (EM) algorithm searching for the maximum likelihood estimators of model parameters. Since the GMM takes into account the covariance structure of the given data as well as the centers of the components, it is considered a generalization of $k$-means clustering. We will use the means and the variances generated by the GMM to create two overlapping intervals for the cover parameter in Mapper.

\begin{figure}[htp]
    \centering
    \begin{subfigure}[b]{0.45\textwidth}
        \centering
        \includegraphics[width=1\textwidth]{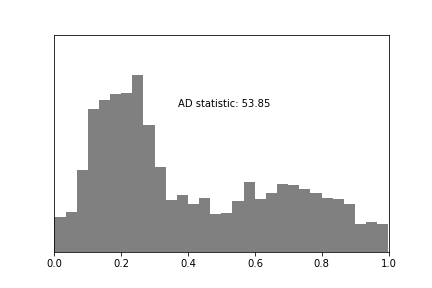}
        \caption{A $1$-dimensional dataset}
        \label{fig:1d_data}
    \end{subfigure}
    \hfil
    \begin{subfigure}[b]{0.45\textwidth}
        \centering
        \includegraphics[width=1\textwidth]{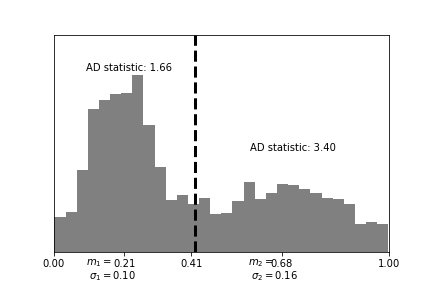}
        \caption{Two subsets of the dataset}
        \label{fig:Two_subsets}
    \end{subfigure}
    \caption{Anderson-Darling statistics and a Gaussian mixture model. \figref{1d_data}. The histogram of a $1$-dimensional dataset whose AD statistic is $53.85$. \figref{Two_subsets}. The GMM is applied to the dataset and the AD statistics of the left and right sides are much smaller than the AD statistic of the entire dataset.}
    \label{fig:AD_GMM}
\end{figure}

\begin{example}[AD Statistic and GMM] \label{ex:AD_GMM}
\normalfont
Let $X$ be a one-dimensional dataset whose histogram is shown in \figref{AD_GMM}. The AD statistic of the dataset $X$ is $53.85$. If we apply a GMM with two components on this dataset $X$, we get means $m_1=0.21$, $m_2=0.68$ and variances $\sigma_1=0.10$, $\sigma_2=0.16$. We divide the dataset into two sets based on if a point has a value less than $0.41$ or not. The AD statistics of the left and right sides are $1.66$ and $3.40$, respectively. These values are much smaller than the AD statistic of the entire dataset $X$. 
\end{example}
\vskip -10pt

\subsection{\texorpdfstring{G}{G}-means clustering} \label{sec:g-means}
The G-means clustering algorithm \cite{hamerly2003learning} automatically determines the number $k$ of clusters by employing the Anderson-Darling test to decide if a cluster should be split into two clusters. The algorithm starts by applying the $k$-means clustering to a dataset of vectors with a small number of centers. The algorithm is initialized with $k=1$, or a larger $k$ can be specified if there is some advance knowledge of the range of values of $k$.

Let $\mathcal{C}$ be the set of derived clusters. For a given cluster $C \in \mathcal{C}$, the algorithm tests whether the cluster $C$ should be split. Applying $k$-means with two components to $C$ splits the cluster into two clusters $C_1$ and $C_2$. Let $\ell$ be the line connecting the two centers of $C_1$ and $C_2$ and project the points of the cluster $C$ onto $\ell$. If the corrected Anderson-Darling statistic for the projected points is below the user-specified AD threshold, the cluster $C$ is kept. Otherwise, the cluster $C$ is removed from $\mathcal{C}$ and $C_1$ and $C_2$ are added instead. These steps are repeated until no more clusters are added.

%% file: body/method.tex
\section{Methods}

We devise a method (G-Mapper) for learning a cover in the Mapper construction based on the G-means algorithm for learning the parameter $k$ in identifying the correct number $k$ of clusters. 

\subsection{\texorpdfstring{G}{G}-Mapper Algorithm}\label{sec:G-Mapper_Algorithm}
The input for G-Mapper consists of (1) the image of a dataset $X$ under the $1$-dimensional lens $f$, i.e., $f(X)$, (2) the AD threshold, and (3) the percentage of overlap for intervals when an interval is split into two. This parameter is denoted by \text{{\fontfamily{cmtt}\selectfont g\_overlap}}. 

The main differences between the G-Mapper algorithm and the G-means clustering algorithm are (1) we are interested in finding overlapping clusters (intervals) rather than disjoint clusters, (2) we design overlapping intervals using variances generated by the GMM rather than applying $k$-means, and (3) the G-Mapper algorithm also does not require a projection of points for performing the Anderson-Darling test since the lens employed in our algorithm is $1$-dimensional.

For G-Mapper, we begin with the cover consisting of one interval containing $f(X)$, i.e., the interval $[\min\{f(X)\}, \max\{f(X)\}]$. The G-Mapper algorithm is iterative and proceeds as follows: 
\begin{enumerate}
\item Select an interval from the current cover (a collection of intervals).

\item For the data points in the interval, perform a statistical test to determine if it follows a Gaussian distribution. For the test, we use the corrected Anderson-Darling statistic. 

\item If the computed statistic is smaller than the AD threshold, keep the original interval.

\item Otherwise, split the interval into two overlapping intervals. We utilize the means and variances derived from the GMM, as described in detail in the following paragraph.

Repeat Steps 1-4 until no more intervals split. 
\end{enumerate}
See \exref{Gmapper} for a step-by-step illustration of the G-Mapper algorithm.

\begin{figure}[htp]
        \centering
        \includegraphics[width=0.9\textwidth]{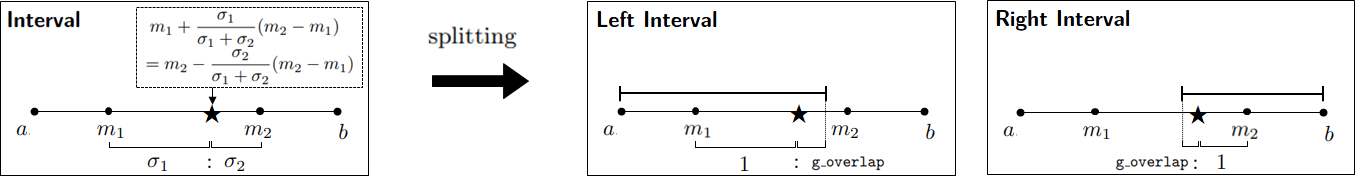}
        \vskip -5pt
    \caption{The point $\bigstar$ divides the two means, $m_1$ and $m_2$, in the ratio $\sigma_1:\sigma_2$, where $\sigma_1$ and $\sigma_2$ are the standard deviations. Two intervals are created by extending the distance between $m_1$ (or $m_2$) and $\bigstar$ by \goverlap\!\!\!, respectively.}
    \label{fig:Splitting}
\end{figure}

We now provide the details of the splitting procedure in Step 4. Let $m_1$, $m_2$ and $\sigma_1$, $\sigma_2$ be the two means and the two standard deviations discovered from the GMM. An interval $(a, b)$ is split into the following two intervals: \vskip -30pt
\begin{align*}
    &(a, \min\{m_1+(1+ \text{{\fontfamily{cmtt}\selectfont g\_overlap}})\sigma_1/(\sigma_1+\sigma_2)(m_2-m_1), m_2\}), \mbox{ and }
\\
&(\max\{m_2-(1+\text{{\fontfamily{cmtt}\selectfont g\_overlap}})\sigma_2/(\sigma_1+\sigma_2)(m_2-m_1), m_1\}, b).
\end{align*}
These two intervals are formed by considering a value \cite{Fesser2023}
\[ 
\dfrac{\sigma_2}{\sigma_1+\sigma_2}m_1+\dfrac{\sigma_1}{\sigma_1+\sigma_2}m_2
=m_1+\dfrac{\sigma_1}{\sigma_1+\sigma_2}(m_2-m_1)
=m_2-\dfrac{\sigma_2}{\sigma_1+\sigma_2}(m_2-m_1)
\]
dividing the two means $m_1$ and $m_2$ in the ratio $\sigma_1:\sigma_2$ is based on the minimum-error decision boundary between two Gaussian distributions \cite{duda2012pattern} and extending the length between this value and each mean by \goverlap\!\!\!, respectively. Refer to \figref{Splitting} for an illustration of the splitting procedure.
\vskip -3pt
\vskip -15pt
When applying the GMM algorithm to an interval, we take as two initial centers $c \pm \sqrt{2\lambda/\pi}$, where $c$ and $\lambda$ are the mean and variance of data points in the interval, respectively. This method is the $1$-dimensional version of a recommended initialization using principal component analysis (PCA) \cite{hamerly2003learning}. However, even though we have a consistent way of initializing the centers, the initialization of the variances and mixing coefficients may be different each time we implement the G-Mapper algorithm on the same dataset. Thus, this can result in different G-Mapper graphs with the same parameters on the dataset.

The splitting algorithm has three variations based on deciding which unprocessed interval to check for Step 1. The following three search methods include:
\begin{itemize}
\item The \textit{depth-first search} (DFS) method iteratively splits as deep as possible by choosing an interval with a bigger Anderson-Darling statistic. Once a chosen interval does not split, the method backtracks and explores an unchecked interval. 
\item The \textit{breadth-first search} (BFS) method checks all intervals in the current cover. The interval with the biggest Anderson-Darling statistic is split. This process is repeated with the newly updated cover. 
\item The \textit{randomized} method selects an interval at random, with intervals having a larger Anderson-Darling statistic being more likely to be chosen.
\end{itemize}
In this paper, we adopted the DFS method for generating G-Mapper graphs, which is more time-efficient than the BFS method and more reproducible than the randomized method.
\begin{example}[The G-Mapper Algorithm]\label{ex:Gmapper}
\normalfont
We apply the G-Mapper algorithm to the circle dataset given in \exref{mapper}. We initialize the algorithm by selecting an AD threshold equal to $10$, \goverlap equal to $0.2$, and the DBSCAN clustering algorithm. The G-Mapper algorithm starts with the cover $\{U_0\}$ consisting of one interval, the entire target space $U_0=Y=[-0.03, 1.05]$. The Mapper graph at this iteration is a single vertex. In the first iteration, the AD statistic of $U_0$ is $32.09$. Since $32.09>10$, the first interval is split into two intervals $U_0'=[-0.03,0.65)$ and $U_0''=(0.52,1.05]$. The Mapper graph at this iteration is an edge. In the second iteration, the AD statistics of $U_0'$ and $U_0''$ are $3.30$ and $10.64$, respectively. Since $10.64>10$, the interval $U_0''$ is split into $U_2=(0.52,0.89)$ and $U_3=(0.84,1.05]$ so that $\{U_1,U_2,U_3\}$ is a cover of $Y$ with $U_1=U_0'=[-0.03,0.65)$. The Mapper graph is a cycle graph with $4$ vertices.
\end{example}
\begin{figure}[htp]
    \centering
    \begin{subfigure}[b]{0.3\textwidth}
        \centering
        \includegraphics[width=1\textwidth]{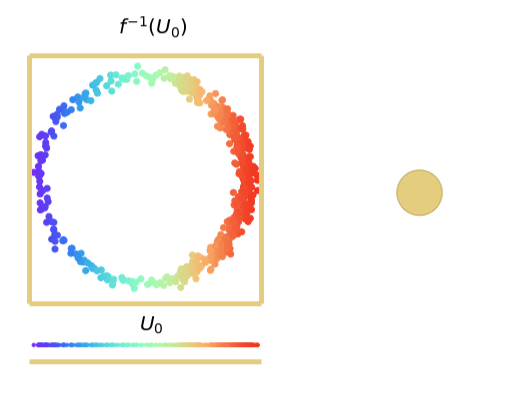}
        \caption{Iteration $0$}
        \label{fig:iter_0}
    \end{subfigure}
    \hfill
    \begin{subfigure}[b]{0.3\textwidth}
        \centering
        \includegraphics[width=1\textwidth]{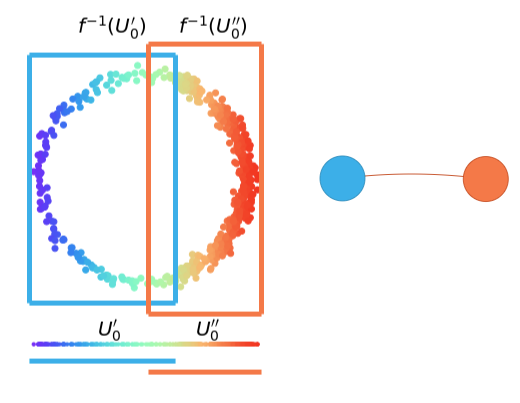}
        \caption{Iteration $1$}
        \label{fig:iter_1}
    \end{subfigure}
        \hfill
    \begin{subfigure}[b]{0.3\textwidth}
        \centering
        \includegraphics[width=1\textwidth]{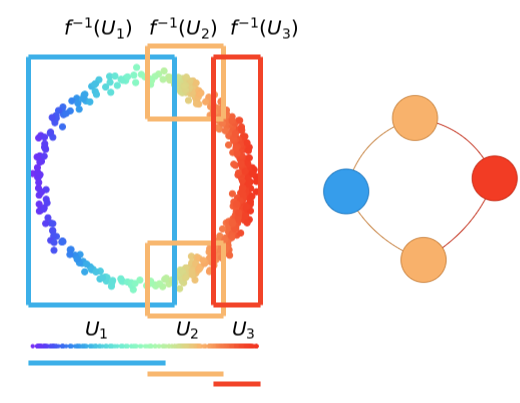}
        \caption{Iteration $2$}
        \label{fig:iter_2}
    \end{subfigure}
    \caption{G-Mapper. The initialization and first two iterations of the splitting procedure are represented in \figref{iter_0}, \figref{iter_1}, and \figref{iter_2} respectively. The cover, the pre-images of the cover elements, and the corresponding Mapper graph are located on the lower left side, the upper left side, and the right side, respectively. The final Mapper graph is a cycle graph with $4$ vertices.}
    \label{fig:GMapper_Ex}
    \vskip -20pt
\end{figure}

\subsection{Parameter Selection for the \texorpdfstring{G}{G}-Mapper Algorithm}
The G-Mapper algorithm involves two key user-specified parameters: \text{{\fontfamily{cmtt}\selectfont AD\_threshold}}, the critical value corresponding to the significance level $\alpha$ for the statistical test, and \text{{\fontfamily{cmtt}\selectfont g\_overlap}}, the amount of overlap when an interval is split in two. 
Note that, although the number of parameters to be chosen remains unchanged, \text{{\fontfamily{cmtt}\selectfont AD\_threshold}} is a statistical measure that is easy to set, taking into account the distribution of the image of a lens function. In addition, unlike the Multipass AIC/BIC algorithm, the G-Mapper algorithm does not require the user to initialize an open cover. In the next section, we show G-Mapper performs well without this initialization.

The first parameter is \text{{\fontfamily{cmtt}\selectfont AD\_threshold}} corresponding to the significance level $\alpha$ for the statistical test. The significance level is the probability of making a Type I error meaning we incorrectly reject the null hypothesis $H_0$. As the parameter (resolution) for the number of intervals of a cover in the conventional Mapper algorithm, \text{{\fontfamily{cmtt}\selectfont AD\_threshold}} influences the number of nodes of the output graph. Setting a lower threshold segments the given data into more pieces, which produces a more detailed Mapper graph with more nodes. Taking a high threshold yields a coarse-grained visualization with the graph having fewer nodes.

The second parameter is \goverlap which specifies how much two intervals should overlap when applying a split. In the G-Mapper algorithm, a split is made utilizing the means and variances estimated by the GMM together with this parameter. As the overlap parameter (gain) in the conventional Mapper algorithm, \goverlap controls relationships between overlapping clusters. Increasing this parameter generates more edges between nodes in the Mapper graph, which results in a more compact graph representation. Decreasing this parameter makes the output Mapper graph less connected, segmenting nodes in the graph into smaller groups.

\begin{figure}[!htb]
    \begin{center}
    \includegraphics[width=.55\textwidth]{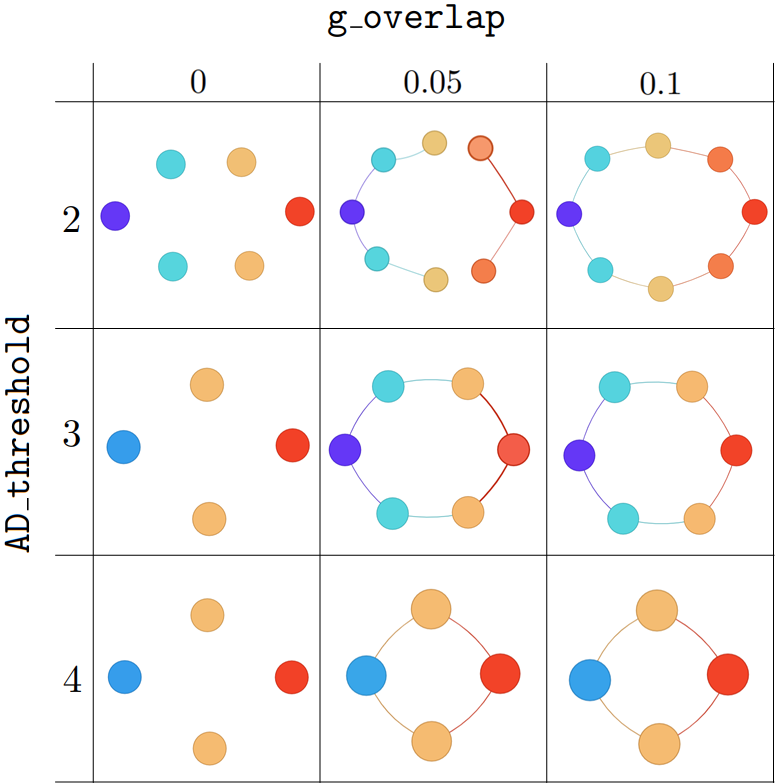}
    \caption{Variation of G-Mapper Parameters for the Circle Dataset. As the $\texttt{g\_overlap}$ parameter increases, the Mapper graph becomes more connected. As $\texttt{AD\_threshold}$ parameter increases, the Mapper graph becomes more coarse.}
    \label{fig:gmapper-params}
    \end{center}
\end{figure}

In~\figref{gmapper-params}, we provide examples of the effects of varying the hyperparameters (\text{{\fontfamily{cmtt}\selectfont AD\_threshold}} 
 and \goverlap\!\!\!) for the circle dataset.

%% file: body/results.tex
\section{Experimental Results}\label{sec:Results}

In this section, we present the results of applying the G-Mapper algorithm to both synthetic and real-world datasets compared to fine-tuned Mapper graphs resulting from the conventional Mapper algorithm. We call the fine-tuned Mapper graphs the \textit{reference Mapper graphs}. This is followed by a comparison to the current state-of-the-art Mapper construction algorithms (Multipass BIC, F-Mapper, and balanced cover). We explore how Mapper graphs built from G-Mapper and Multipass BIC are different. For the latter two algorithms, we utilize the number of intervals detected by G-Mapper as an input parameter. Finally, we close the section by providing a runtime comparison between all methods.
 
\subsection{Synthetic Datasets}\label{sec:Synthetic}

We first applied the G-Mapper algorithm to three synthetic datasets (two circles, human, and Klein bottle datasets). Our experiments reveal that G-Mapper generates Mapper graphs that aptly describe the intrinsic structure of the original datasets. Furthermore, the output G-Mapper graphs turn out to be close to the reference Mapper graphs. For each dataset, three figures for the original dataset and these two Mapper graphs will be presented together.

We describe the two main parameters (\goverlap and \text{{\fontfamily{cmtt}\selectfont AD\_threshold}}) of G-Mapper for the synthetic datasets. For all datasets, we set the parameter \goverlap to $0.1$. We pick different Anderson-Darling (AD) thresholds according to the datasets. We use $10$ as the thresholds for the two circles and human datasets, and a larger value of $15$ for the Klein bottle dataset to prevent generating an overly detailed Mapper graph. 

In order to better visualize the characteristics of nodes in Mapper graphs, we color nodes with the rainbow colormap. In all Mapper graphs of synthetic datasets, the color of a node represents the average value of the lens function over all data points in the node.

\subsubsection{Two Circles Dataset}
\begin{figure}[!htb]
    \centering
    \begin{subfigure}[b]{0.32\textwidth}
        \centering
        \includegraphics[width=.8\textwidth]{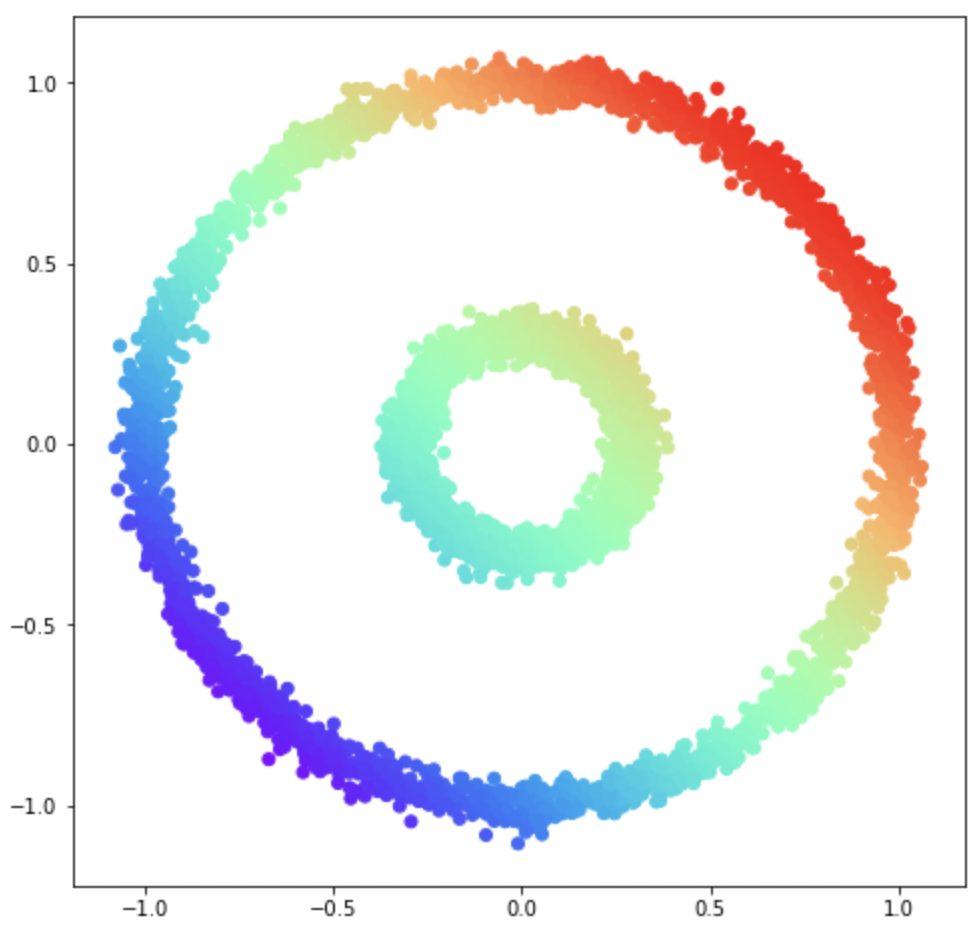}
        \caption{Two Circles Dataset}
        \label{fig:two-circle-data}
    \end{subfigure}
    \hfil
    \begin{subfigure}[b]{0.32\textwidth}
        \centering
        \includegraphics[width=.8\textwidth]{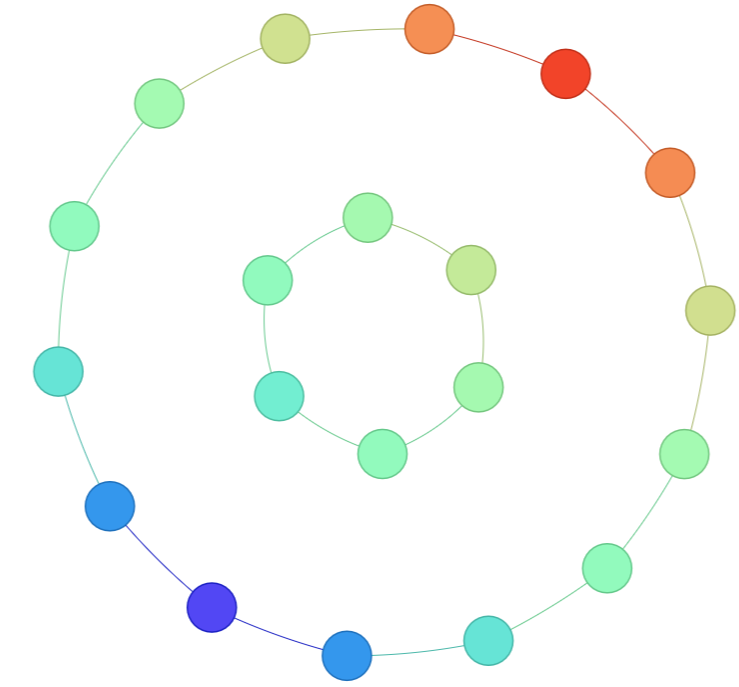}
        \caption{G-Mapper Graph}
        \label{fig:two-circle-dfs}
    \end{subfigure}
    \hfil
    \begin{subfigure}[b]{0.32\textwidth}
        \centering
        \includegraphics[width=.75\textwidth]{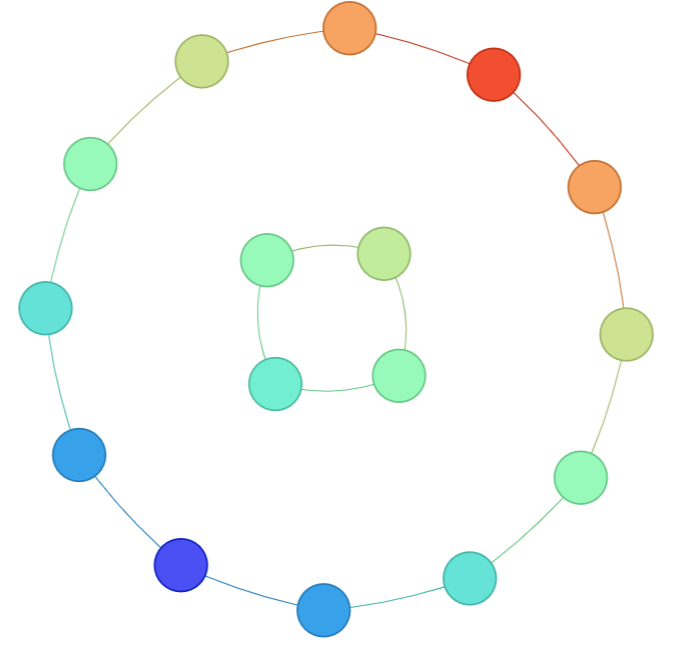}
        \caption{Reference Mapper Graph}
        \label{fig:two-circle-refer}
    \end{subfigure}
    \caption{Two Circles Dataset. G-Mapper Parameters: AD threshold = 10, \goverlap = 0.1, clustering algorithm = DBSCAN with $\varepsilon=0.1$ and \text{{\fontfamily{cmtt}\selectfont MinPts}} = 5, and search method = DFS. The cover was found in 7 iterations and consists of 8 intervals. Reference Mapper Parameters: number of intervals = 7, overlap = 0.2, and the same DBSCAN parameters.}
    \label{fig:two-circles-result}
\end{figure}

The two circles dataset comprises 5,000 sampled points from two concentric circles. The lens function is the sum of the $x$ and $y$ coordinates normalized. The reference Mapper graph consists of two concentric circles. The G-Mapper graph also consists of two concentric circles and the graph was found in 7 iterations. Note that the generated Mapper graphs retain the homological features of the underlying space. In \figref{two-circles-result}, we present Mapper graphs along with the parameters used.
\vskip 10pt

\subsubsection{Human Dataset}

\begin{figure}[!htb]
    \centering
    \begin{subfigure}[b]{0.32\textwidth}
        \centering
        \includegraphics[width=.4\textwidth]{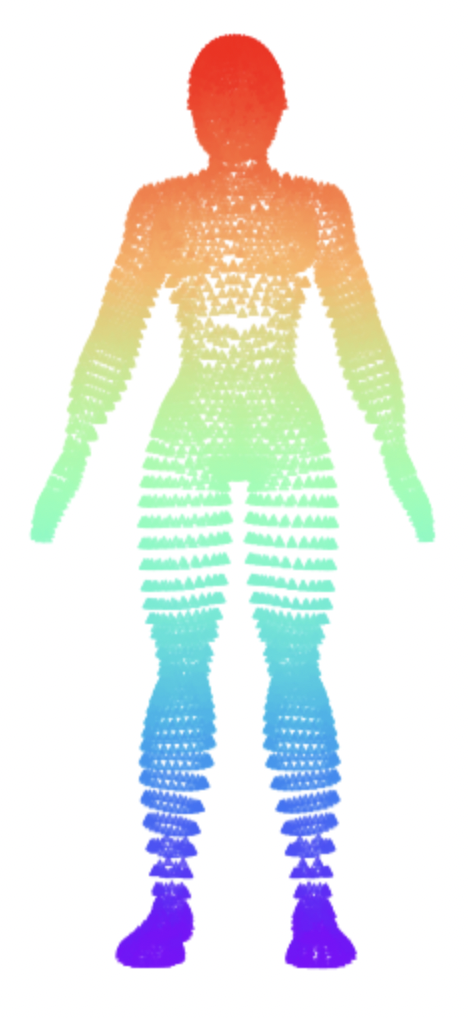}
        \caption{Human Dataset}
        \label{fig:human-data}
    \end{subfigure}
    \hfil
    \begin{subfigure}[b]{0.32\textwidth}
        \centering
        \includegraphics[width=.6\textwidth]{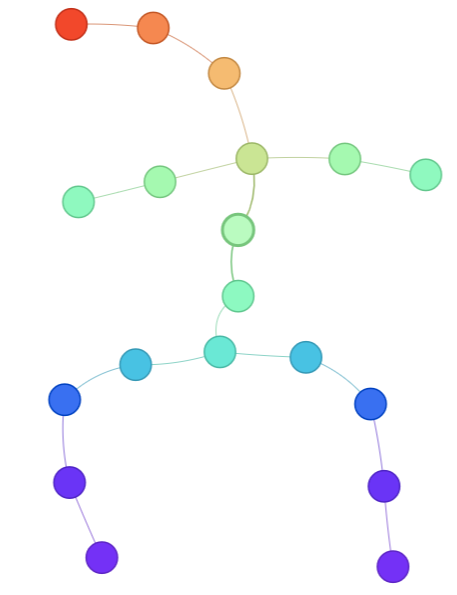}
        \caption{G-Mapper Graph}
        \label{fig:human-dfs}
    \end{subfigure}
    \hfil
    \begin{subfigure}[b]{0.32\textwidth}
        \centering
        \includegraphics[width=.6\textwidth]{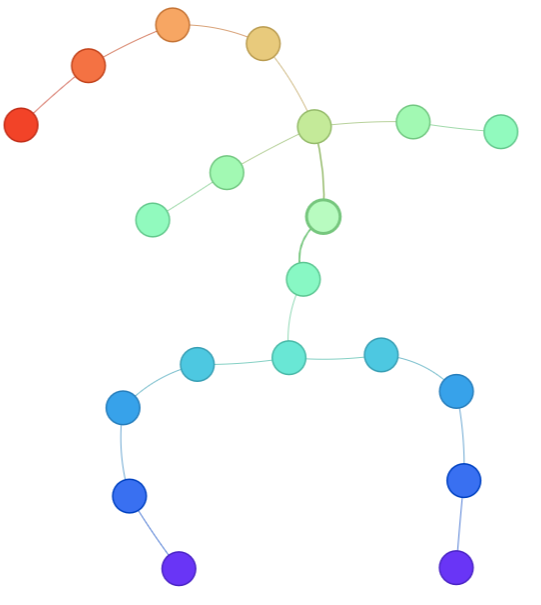}
        \caption{Reference Mapper Graph}
        \label{fig:human-refer}
    \end{subfigure}
    \caption{Human Dataset. G-Mapper Parameters: AD threshold = 10, 
\goverlap = 0.04, clustering algorithm = DBSCAN with $\varepsilon = 0.1$ and \text{{\fontfamily{cmtt}\selectfont MinPts}} = 5, and search method = DFS. The cover was found in 10 iterations and consists of 11 intervals. Reference Mapper Parameters: number of intervals = 12, overlap = 0.2, and the same DBSCAN parameters.
}
    \label{fig:human-result}
\end{figure}

The next point cloud dataset we explored is a 3D human shape from \cite{chen2009benchmark} that consists of 4,706 points. The lens function is the normalized height function. In \figref{human-result}, we present Mapper graphs along with the parameters used. The cover for G-Mapper was found in $10$ iterations. Both the reference Mapper graph and the G-Mapper graph represent human skeletons consisting of a head, two arms, and two legs. Two Mapper graphs have identical except for the number of nodes in the head and the coloring of nodes in the legs.
\vskip 10pt

\subsubsection{Klein Bottle Dataset}

\begin{figure}[!htb]
    \centering
    \begin{subfigure}[b]{0.32\textwidth}
        \centering
        \includegraphics[width=.8\textwidth]{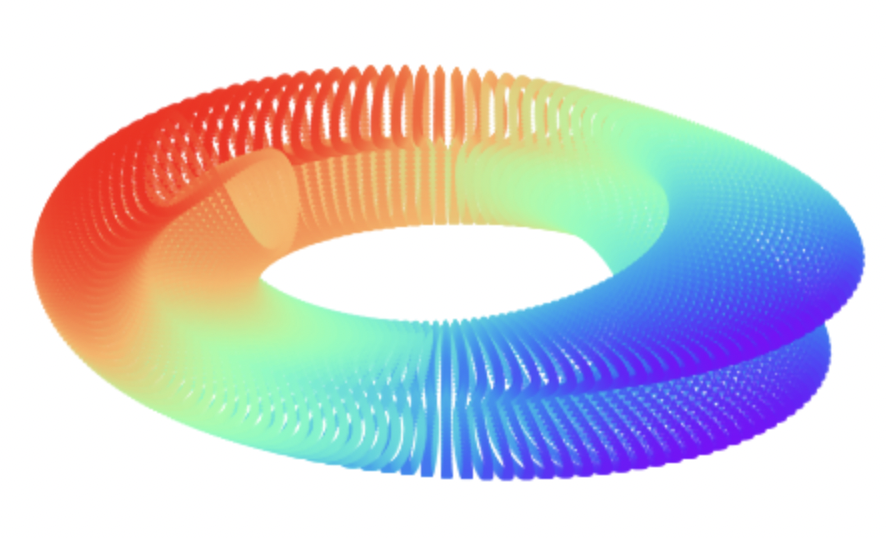}
        \caption{3D Projection}
        \label{fig:klein-bottle-data}
    \end{subfigure}
    \hfil
    \begin{subfigure}[b]{0.32\textwidth}
        \centering
        \includegraphics[width=.7\textwidth]{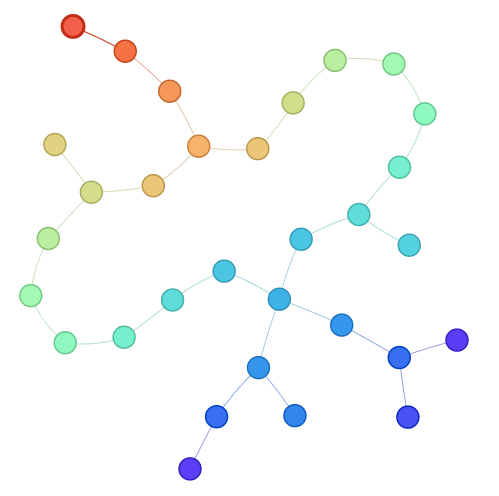}
        \caption{G-Mapper Graph}
        \label{fig:klein-bottle-dfs}
    \end{subfigure}
    \hfil
    \begin{subfigure}[b]{0.32\textwidth}
        \centering
       \includegraphics[width=.7\textwidth]{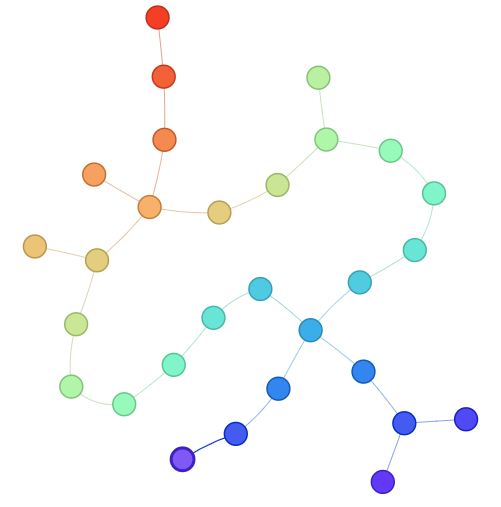}
        \caption{Reference Mapper Graph}
       \label{fig:klein-bottle-refer}
    \end{subfigure}
    \caption{Klein Bottle Dataset. G-Mapper Parameters: AD threshold = 15, \goverlap = 0.05, clustering algorithm = DBSCAN with $\varepsilon=0.21$ and \text{{\fontfamily{cmtt}\selectfont MinPts}} = 5. The cover was found in 16 iterations and consists of 17 intervals. Reference Mapper Parameters: number of intervals = 15, overlap = 0.2, and the same DBSCAN parameters.}
    \label{fig:klein-bottle-result}
\end{figure}

The Klein bottle dataset consists of 15,875 points sampled from the Klein Bottle embedded in $\R^5$. The dataset is obtained from the Gudhi library \cite{clement2014gudhi}. The lens function is the projection map onto the first coordinate normalized. In \figref{klein-bottle-result}, we present Mapper graphs along with the parameters used. The reference Mapper graph exhibits a long cycle with several flares (branches). The G-Mapper algorithm produces a highly similar graph in 16 iterations.
\vskip 10pt

\vskip 15pt
\subsection{Real-World Datasets}\label{sec:Real-World}
We next applied G-Mapper to three real-world datasets: Passiflora leaves, COVID-19 trends, and the CIFAR-10 image dataset. We explain these datasets in detail in \secref{passiflora},  \secref{covid}, and \secref{CIFAR-10}, respectively. Each data point is labeled by its morphotype, the state in which it was collected, and its image class in the Passiflora dataset, COVID-19 dataset, and CIFAR-10 image dataset, respectively. These datasets are of higher dimension than the synthetic datasets explored in the previous section. We visualize the nodes in the Mapper graphs as pie charts in order to represent the proportions of data points within the nodes belonging to different labels.
\vskip 25pt
\subsubsection{Passiflora Dataset} \label{sec:passiflora}

The Passiflora dataset \cite{morphology} consists of 3,319 leaves from 40 different species of the Passiflora genus. Leaves of the Passiflora genus are of particular interest to biologists due to their remarkable diversity of shape. Each leaf in the dataset has $15$ landmarks whose locations are $2$-dimensional vectors expressed as $x$ and $y$ coordinates. Consequently, each leaf is represented as a $30$-dimensional vector, and the correlation distance is used to measure the distance between leaf vectors. The correlation distance between two vectors is defined as $1-r$, where $r$ is the Pearson correlation coefficient. In this metric, leaf vectors with a high correlation will have a distance near zero.

\begin{figure}[!htb]
    \centering
    \begin{subfigure}[b]{0.32\textwidth}
        \centering
       \includegraphics[width=1\textwidth]{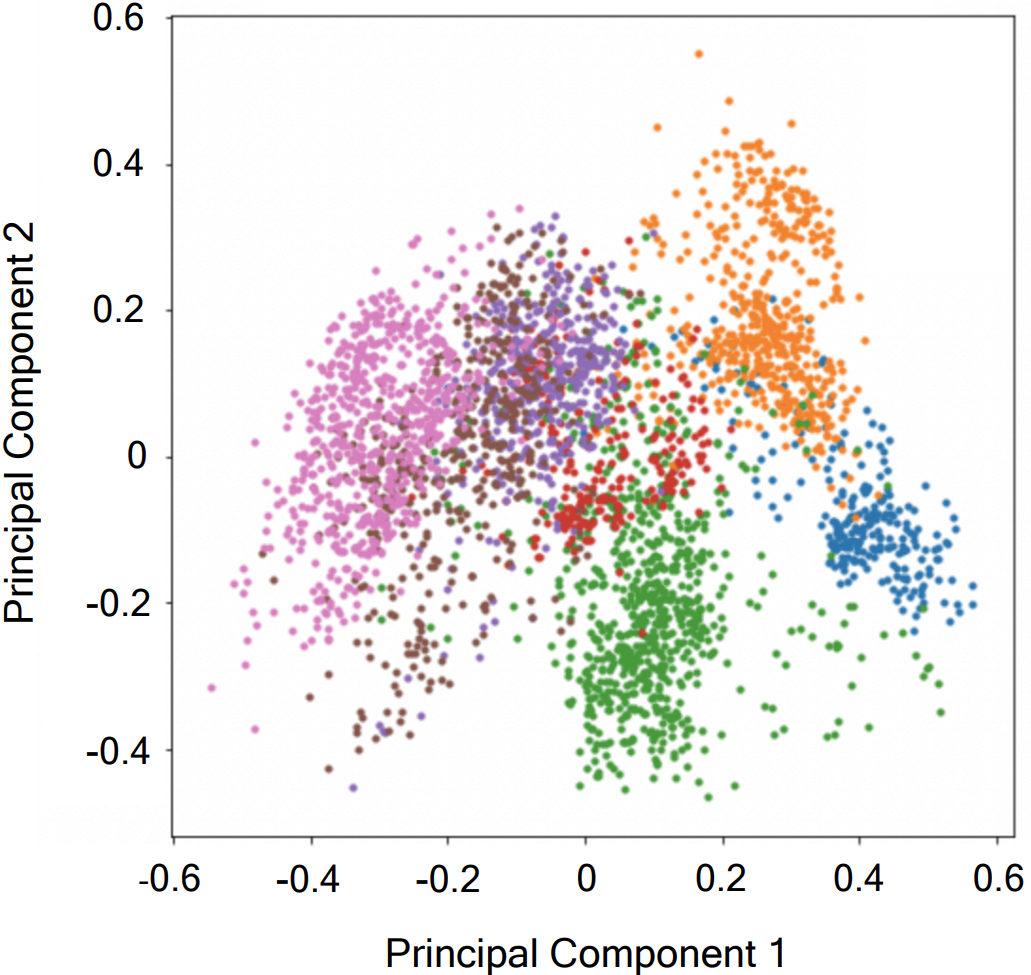}
        \caption{Passiflora PCA Plot}
        \label{fig:PCA-passiflora}
    \end{subfigure}
    \hfil
    \begin{subfigure}[b]{0.32\textwidth}
        \centering
        \includegraphics[width=.95\textwidth]{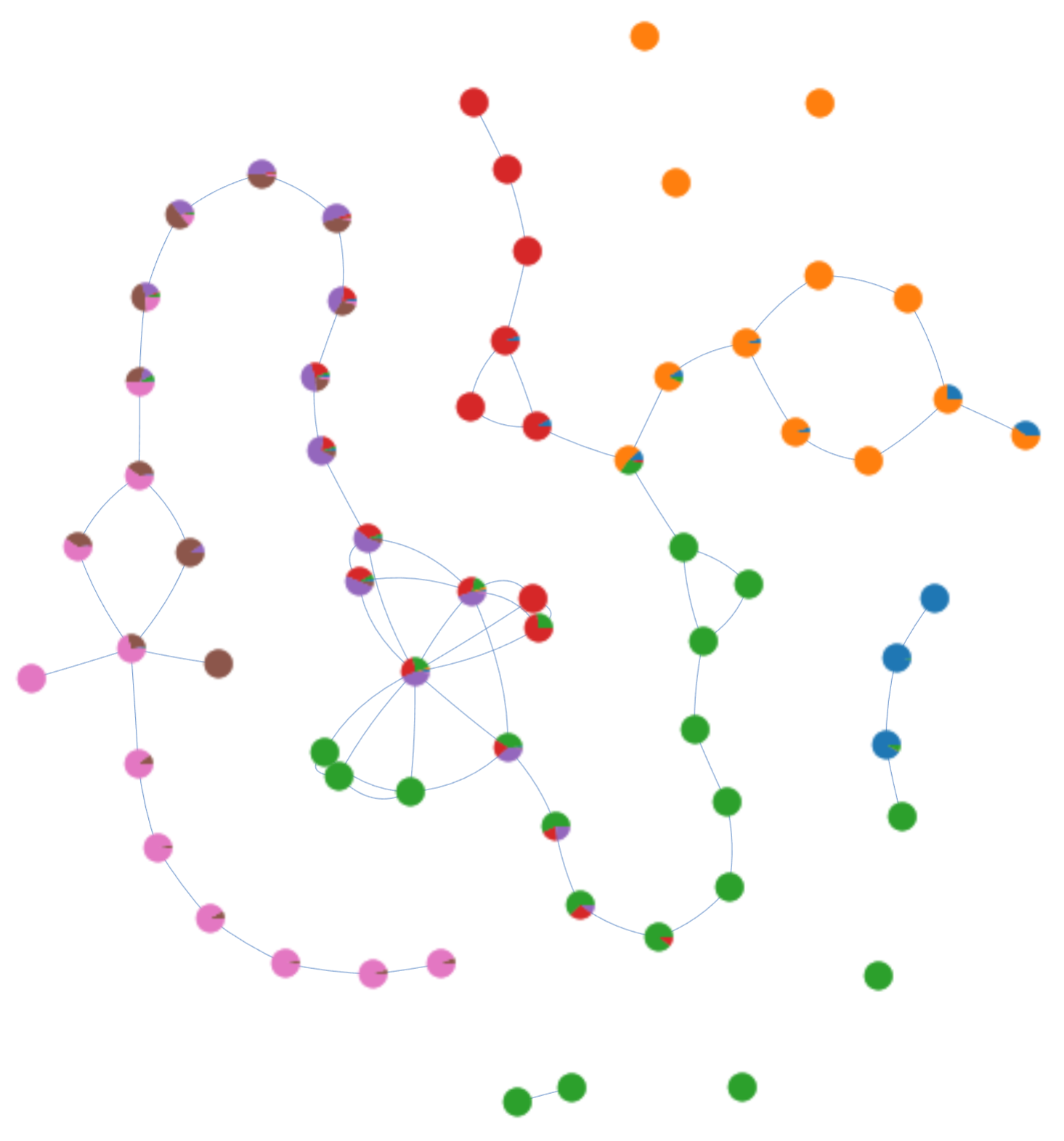}
        \caption{G-Mapper Graph}
        \label{fig:PCA-passiflora-dfs}
    \end{subfigure}
    \hfil 
    \begin{subfigure}[b]{0.32\textwidth}
        \centering
       \includegraphics[width=\textwidth]{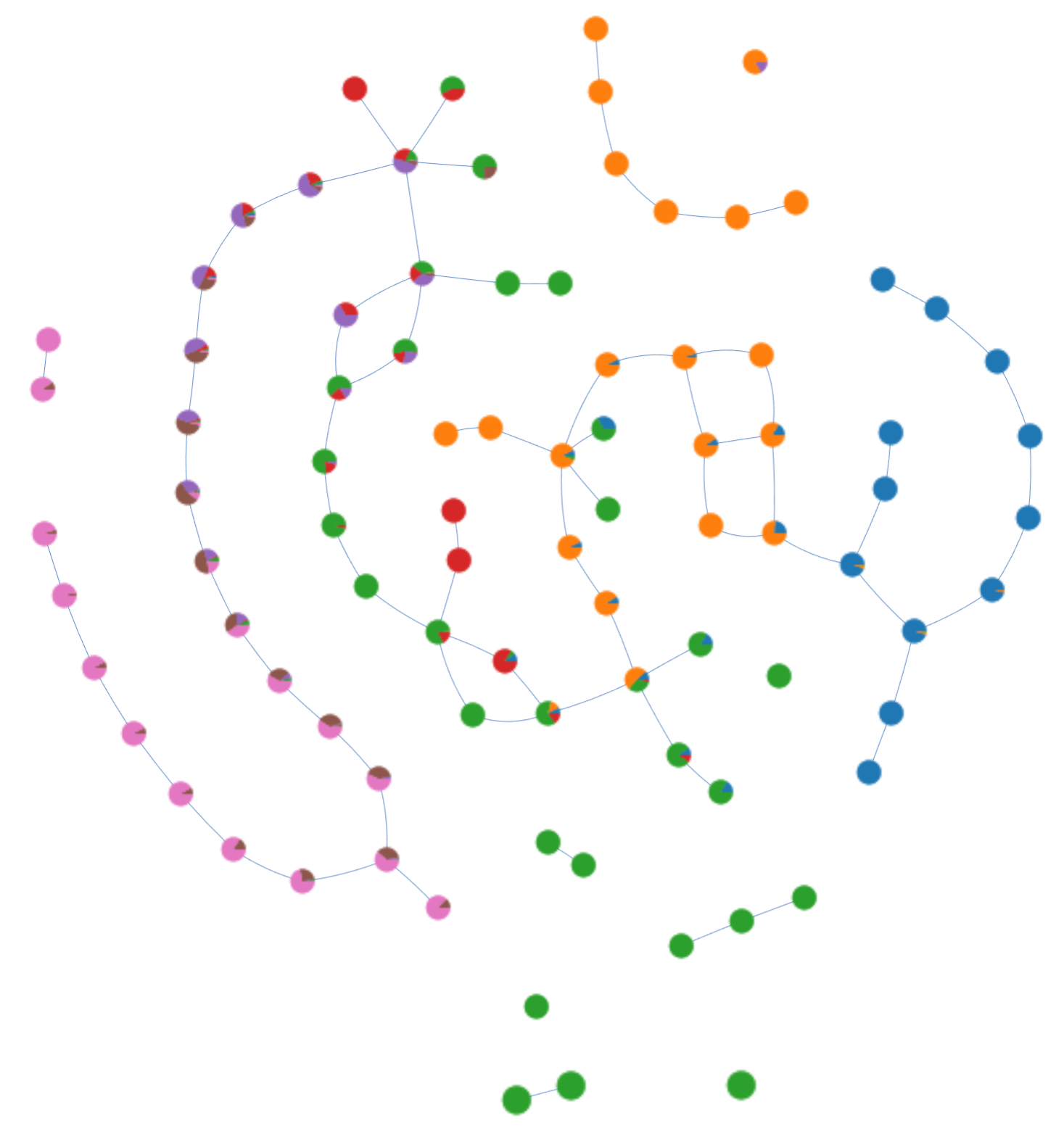}
        \caption{Reference Mapper Graph}
        \label{fig:PCA-passiflora-refer}
    \end{subfigure}
\begin{subfigure}[b]{0.38\textwidth}
    \centering
    \includegraphics[width=\textwidth]{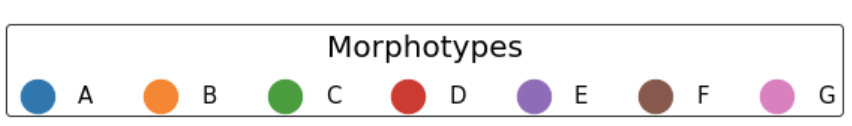}
\end{subfigure}

    \caption{Passiflora Dataset. Lens Function: the first principal component normalized. G-Mapper Parameters: AD threshold = 2, \goverlap = 0.1, clustering algorithm = DBSCAN with $\varepsilon = 0.15$, \text{{\fontfamily{cmtt}\selectfont MinPts}} = 5, and distance = correlation distance, and search method = DFS. The open cover was found in 37 iterations and consists of 38 intervals. Reference Mapper Parameters: number of intervals = 40, overlap = 0.5, and the same DBSCAN parameters.}
    \label{fig:passiflora-pca}
\end{figure}

The authors of \cite{morphology} classified the $40$ species into seven different morphotypes. For the classification, they performed principal component analysis (PCA) on the landmark dataset described in the previous paragraph and elliptical Fourier descriptors on the outline of the leaf. The first and second principal components are visualized in \figref{PCA-passiflora} (given in \cite[Figure 5]{morphology} and \cite[Figure 3C]{percival2024topological}). Because of the significant amount of overlap shown in the PCA plot, the authors also relied on their domain knowledge when assigning morphotypes. Mapper may help extend this process further to obtain better separation between morphotypes and extract hidden relationships between morphotypes. The Mapper algorithm has been utilized in \cite{percival2024topological} for this purpose. 

In \figref{PCA-passiflora-dfs}, we present the results of applying G-Mapper to the Passiflora dataset using the first principal component as the lens function. 
 The G-Mapper graph has a strong linear backbone along the first principal component. The Mapper graph has multi-colored nodes with purple, brown, and red, meaning there is much overlap in the morphotypes, whereas it has single-colored nodes with orange, green, and pink, meaning the morphotypes are more distinct, as shown in the PCA plot. The reference Mapper graph shown in \figref{PCA-passiflora-refer} is generated based on parameters given in \cite{percival2024topological}. The G-Mapper graph is close to the reference Mapper graph in that it has a strong linear backbone and the coloring of nodes is similar. 
 
 However, these two Mapper graphs have different features originating from their constructions. The G-Mapper graph produces more edges between the multicolored purple, red, and green nodes compared to the reference Mapper graph. We suspect this is due to the greater density and overlap of morphotypes having a projection onto the first principal component range between 0-0.2. G-Mapper constructs more intervals in areas that are more dense whereas conventional Mapper constructs intervals in a uniform manner. In addition, the G-Mapper graph has more red nodes and fewer blue nodes than the reference Mapper graph since G-Mapper takes the distribution of the data into account.

\subsubsection{COVID-19 Dataset} \label{sec:covid}

COVID-19 data was collected in \cite{dong2020interactive} and can be found in the data repository (\url{https://github.com/CSSEGISandData/COVID-19/}). In order to analyze COVID-19 state-wide trends, the authors of \cite{zhou2021mapper} selected 1,431 daily records of COVID-19 cases during 159 days between April 12, 2020 to September 18, 2020 for nine states (AZ, CA, FL, GA, IL, NC, NJ, NY, TX) from the dataset. These nine states were chosen since they had the largest number of confirmed cases. For each day, the dataset has the following $7$ attributes: the numbers of confirmed cases, death cases, active cases, people tested, testing rate, mortality, and the numbers of cases per 100,000 people, where the number of active cases is defined by subtracting the number of death cases and recovered cases from the number of confirmed cases. Since the number of recovered cases is unavailable in some states, we simply estimate the number of active cases by the number of confirmed cases minus the number of death cases.

\begin{figure}[!htb]
    \centering
    \begin{subfigure}[b]{0.32\textwidth}
        \centering
        \includegraphics[width=\textwidth]{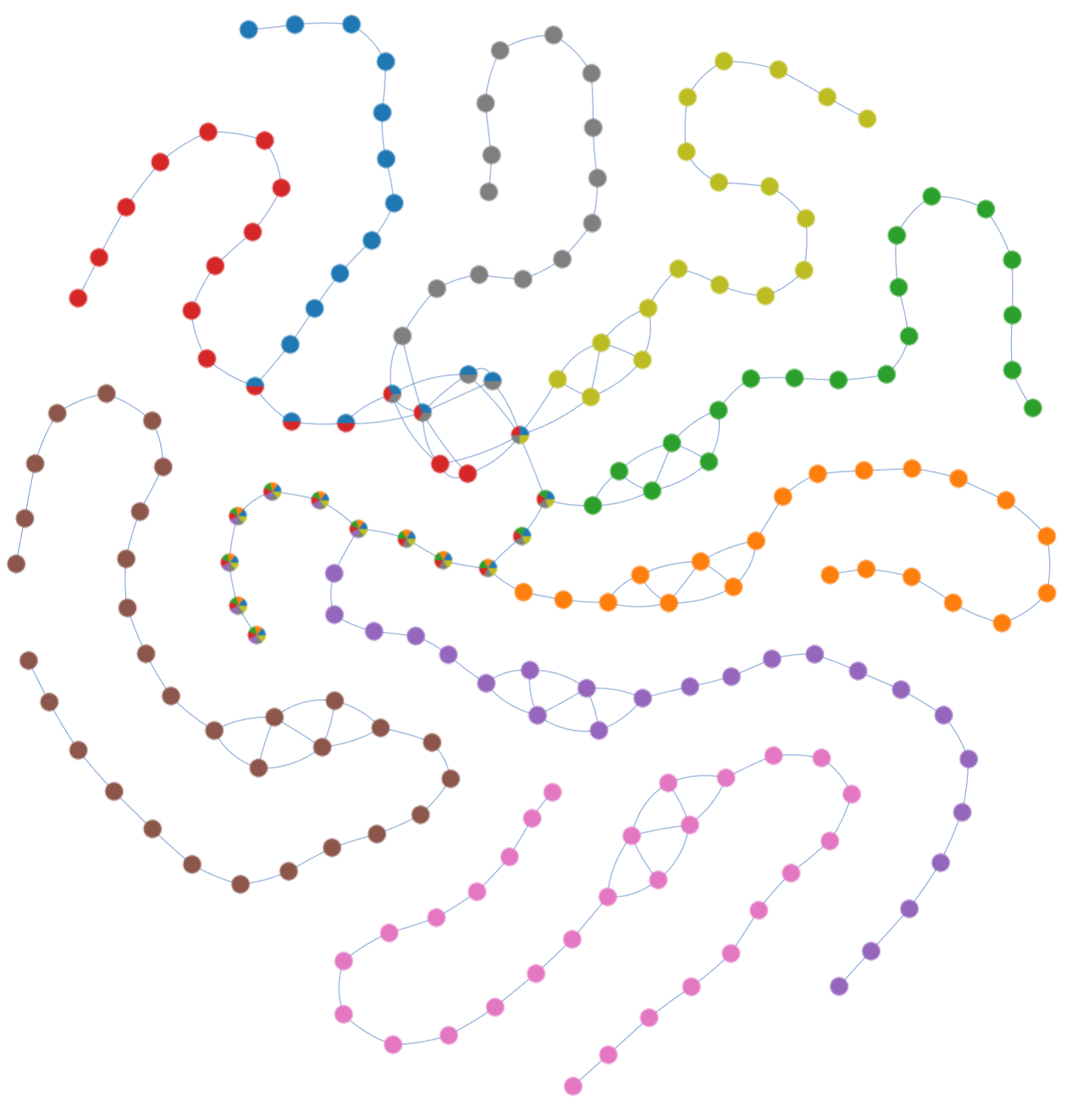}
        \caption{G-Mapper Graph}
        \label{fig:covid-gmapper}
    \end{subfigure}
         \hfil
    \begin{subfigure}[b]{0.32\textwidth}
       \centering
        \includegraphics[width=\textwidth]{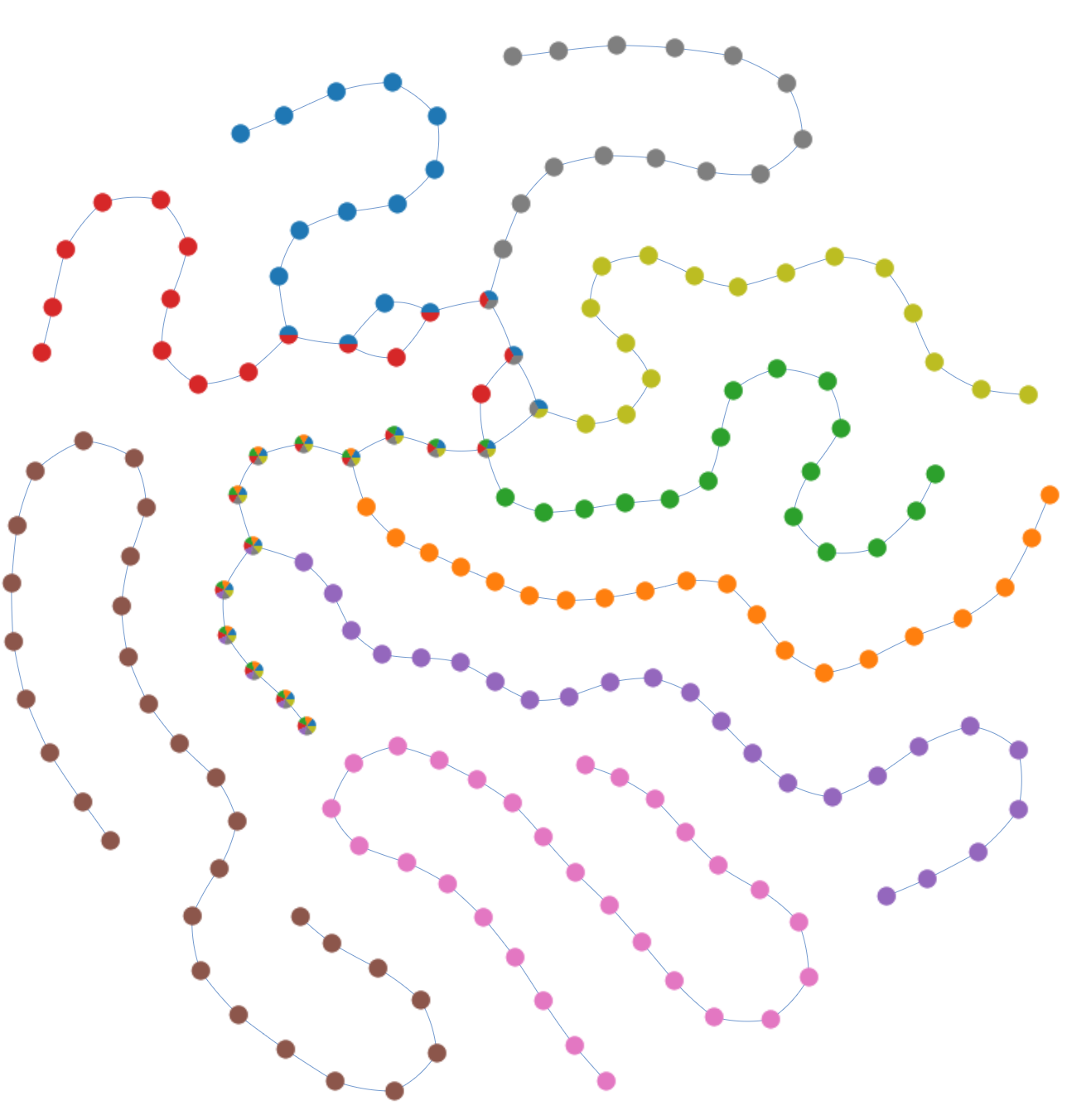}
        \caption{Reference Mapper Graph}
        \label{fig:covid-refer}
    \end{subfigure}
     \hfil
     
    \begin{subfigure}[b]{0.65\textwidth}
              \includegraphics[width=\textwidth]{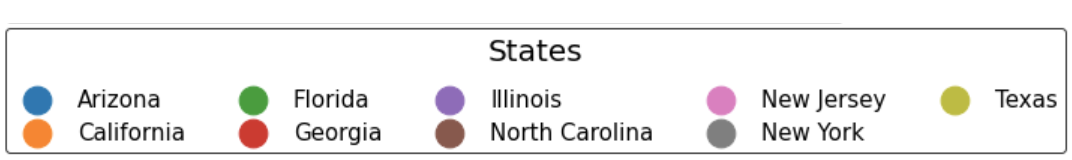}
            \end{subfigure}
            \vskip 10pt
     \caption{COVID-19 Dataset. Lens Function: the number of recorded days normalized. G-Mapper Parameters: AD threshold = 1.35, \goverlap = 0.15, clustering algorithm = DBSCAN with $\varepsilon = 0.15$ and \text{{\fontfamily{cmtt}\selectfont MinPts}}= 5, and search method = DFS. The open cover was found in 30 iterations and consists of 31 intervals. Reference Mapper Parameters: number of intervals = 30, overlap = 0.3, and the same DBSCAN parameters.}
    \label{fig:covid}
\end{figure}

In \figref{covid-gmapper}, we present the results for applying G-Mapper to the COVID-19 dataset along with the parameters. The figure suggests that G-Mapper is able to identify COVID-19 trends according to each state and provides information concerning the relationships between them. The Mapper graph consists of three connected components: the main component, the pink component (NY), and the brown component (NJ), which suggests that NY and NJ have COVID-19 trends distinct from the other states. The purple branch (IL), orange branch (CA), green branch (FL), olive branch (TX), and gray branch (NC) appear sequentially in the main component. Two color nodes with blue (AZ) and red (GA) show up, which indicates that these two states share very similar COVID-19 trends at some point. These two states eventually bifurcate into a blue branch (AZ) and red branch (GA). We observe that the G-Mapper graph closely resembles the reference Mapper graph presented in \figref{covid-refer} since the features listed in this paragraph also appear in the reference Mapper graph.

\subsubsection{CIFAR-10 Dataset} \label{sec:CIFAR-10}
The last real-world dataset we analyzed is a famous image dataset called CIFAR-10 \cite{krizhevsky2009learning}. The dataset consists of 60,000 images (50,000 training images and 10,000 test images) in 10 classes. The 10 different classes represent airplanes, automobiles, birds, cats, deer, dogs, frogs, horses, ships, and trucks. The raw dataset consists of $32\times32$ color images. The input data for applying the Mapper algorithm is obtained by learning the training images with the ResNet-18 neural network, passing the test images through the network, and collecting activation vectors from the last layer. Then 10,000 activation vectors are collected with 1,000 vectors per class, and each vector is $512$-dimensional, which is tremendously higher than the previous two datasets.

\begin{figure}[!htb]
    \centering
    \begin{subfigure}[b]{0.32\textwidth}
        \centering
       \includegraphics[width=\textwidth]{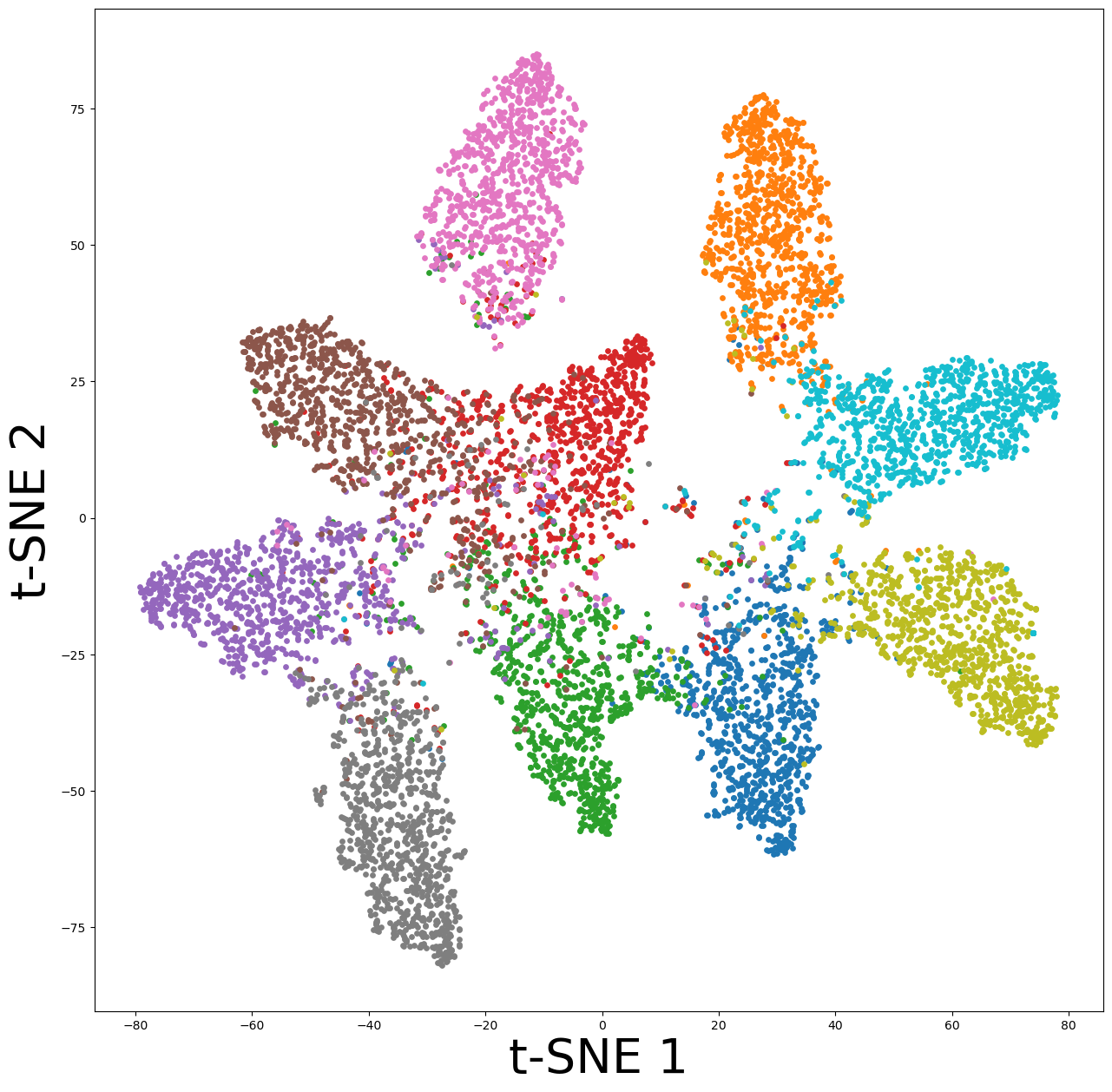}
        \caption{CIFAR-10 t-SNE Plot}
        \label{fig:tSNE-CIFAR-10}
    \end{subfigure}
    \hfil
    \begin{subfigure}[b]{0.32\textwidth}
        \centering
        \includegraphics[width=\textwidth]{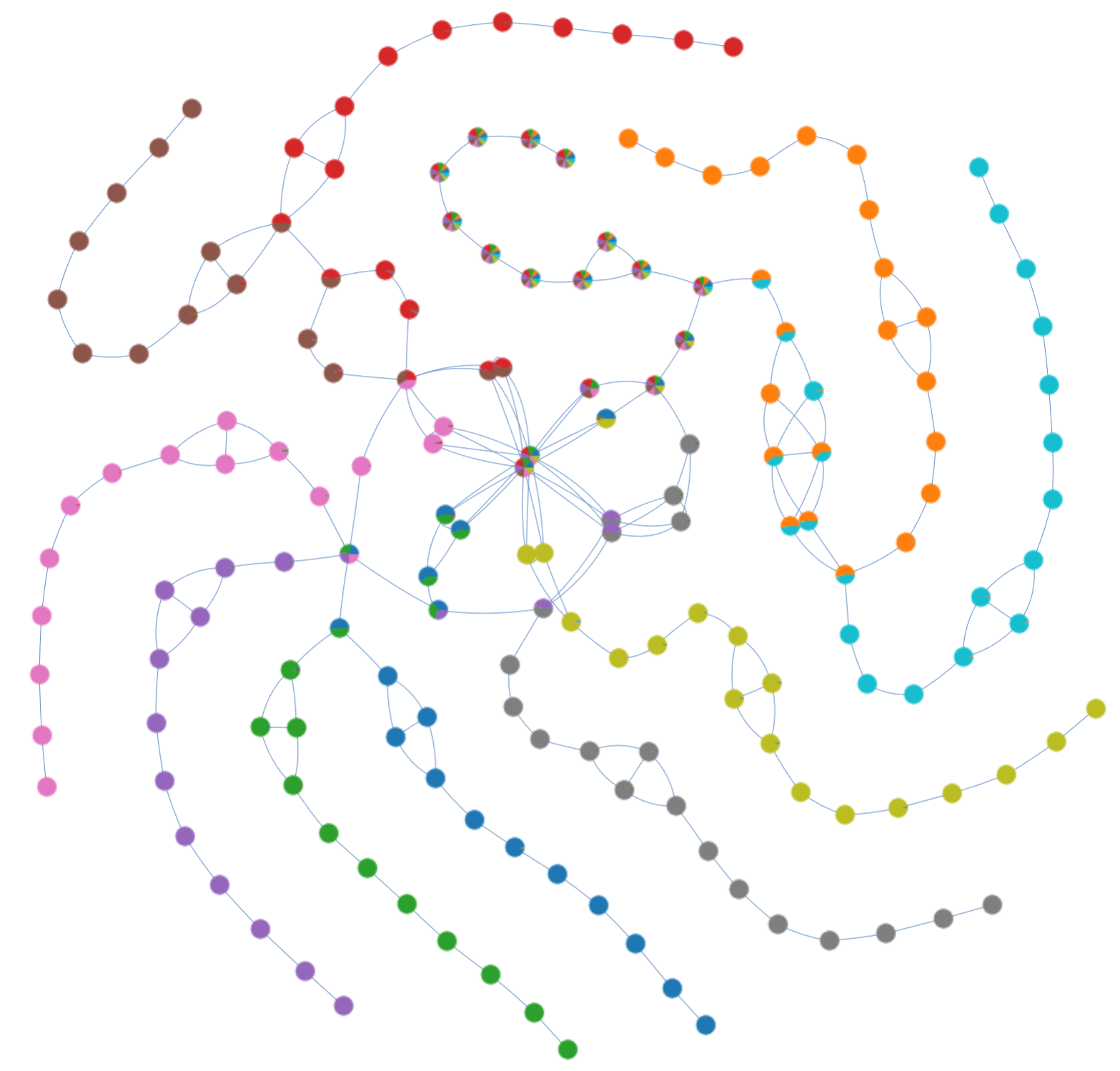}
        \caption{G-Mapper Graph}
        \label{fig:CIFAR-10-dfs}
    \end{subfigure}
    \hfil
    \begin{subfigure}[b]{0.32\textwidth}
        \centering
       \includegraphics[width=\textwidth]{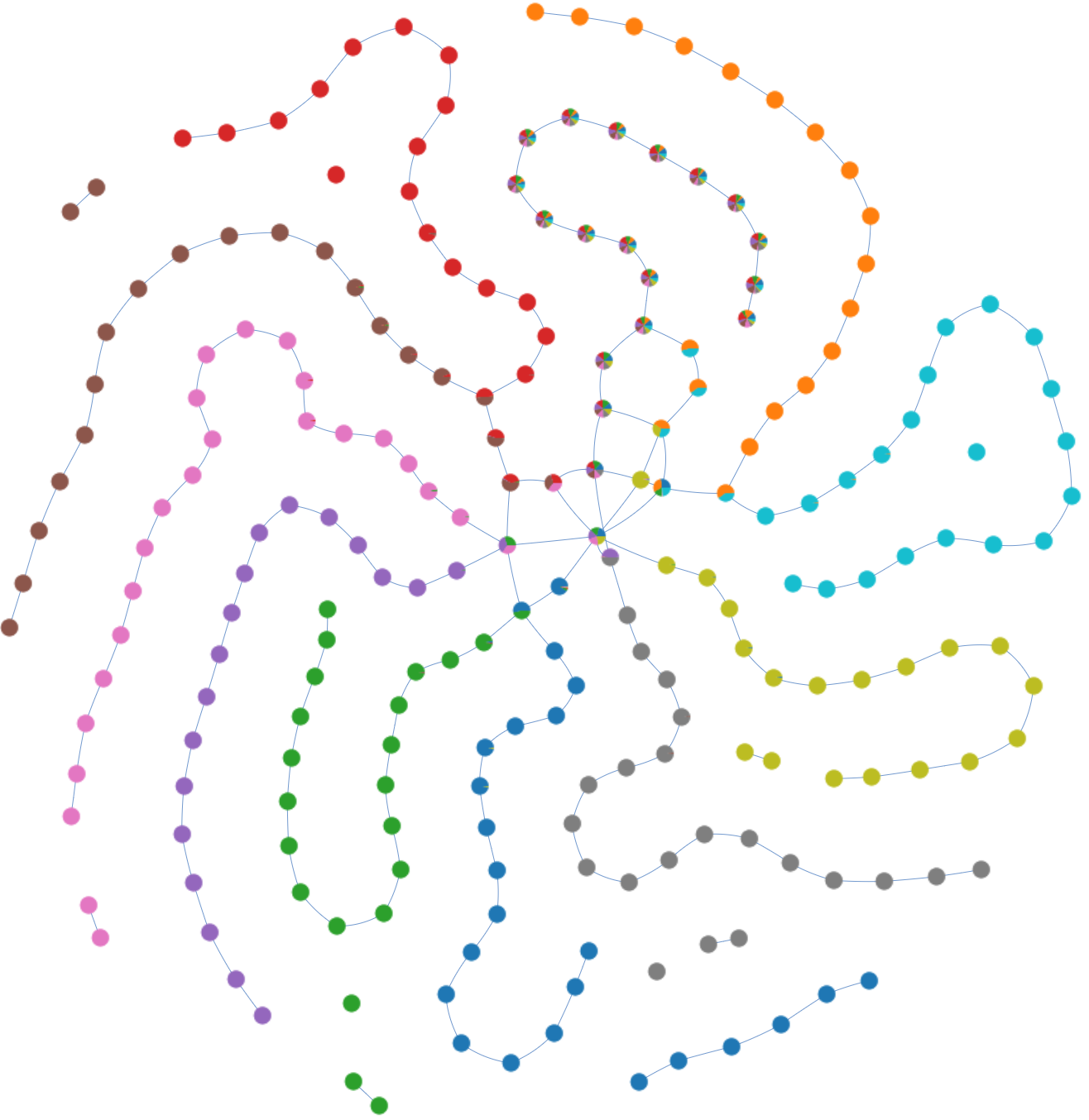}
        \caption{Reference Mapper Graph}
        \label{fig:CIFAR-10-refer}
    \end{subfigure}

    \begin{subfigure}[b]{0.5\textwidth}
    \includegraphics[width=\textwidth]{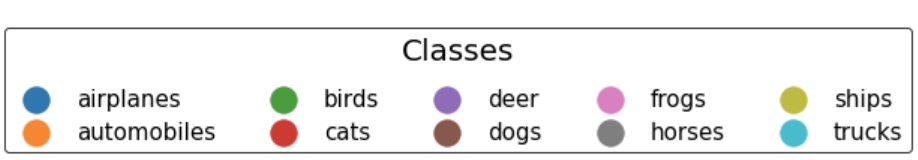}
    \end{subfigure}

    \caption{CIFAR-10 Dataset. Lens Function: the $L_2$ norm of each activation vector. G-Mapper Parameters: AD threshold = 9, \goverlap = 0.2, clustering algorithm = DBSCAN with $\varepsilon = 2$ and \text{{\fontfamily{cmtt}\selectfont MinPts}}$=5$, and search method = DFS. The open cover was found in $32$ iterations and consists of $33$ intervals. Reference Mapper Parameters: number of intervals = 70, overlap = 0.35, and the same DBSCAN parameters.}
    \label{fig:CIFAR-10}

\end{figure}

The t-distributed stochastic neighbor embedding (\emph{t-SNE}) algorithm is a well-known dimensionality reduction method \cite{van2008visualizing}. A 2-dimensional t-SNE embedding of the collection of activation vectors is represented in \figref{tSNE-CIFAR-10}. The plot shows that t-SNE separates the data points into the 10 different classes with some of the classes overlapping. In order to highlight relationships among the 10 classes, \cite{rathore2021topoact, zhou2021mapper, chalapathi2021adaptive} utilized the Mapper algorithm.

\figref{CIFAR-10-dfs} and \figref{CIFAR-10-refer} represent the G-Mapper graph and the reference Mapper graph for the dataset, respectively. Both Mapper graphs classify the data points into the 10 classes and detect three pairs of classes with some similarities. The three pairs are automobiles (orange points) and trucks (cyan points), cats (red points) and dogs (brown points), and airplanes (blue points) and birds (green points). The two classes in each of these three pairs are regarded as having similar features. Each pair has nodes containing data points in both classes, and it eventually bifurcates into two different branches with each branch representing a distinct class.

\subsection{Comparison to Other Methods}
We compare the G-Mapper algorithm to other state-of-the-art techniques on the same synthetic and real-world datasets as above. We analyze how Mapper graphs generated by another iterative algorithm, the Multipass BIC algorithm, are different from G-Mapper graphs. In addition, we produce Mapper graphs from F-Mapper and the balanced cover strategy utilizing the number of cover intervals estimated by G-Mapper as an input parameter. We refer the reader to \cite[Sect. VI-A]{chalapathi2021adaptive} for a discussion and analysis of the performance of the statistical cover strategy. Throughout this section, we maintain the same DBSCAN parameters as used in G-Mapper for each dataset.

\subsubsection{Multipass BIC}\label{sec:Multipass}
The Multipass BIC algorithm repeatedly splits intervals uniformly from some initialized coarse cover based on information criteria. The main parameters that need to be specified for this algorithm include the initial number of intervals used before splitting and the amount of overlap between consecutive intervals in the cover. There is also a threshold parameter $\delta$ for deciding when to split, but this is often set to zero meaning that a split is performed as long as the information criterion statistic improves. We set $\delta=0$, use the BIC statistic, and use the DFS method. We tried to follow the parameters specified in \cite{chalapathi2021adaptive}, but we had to adjust overlap parameters due to an error in the Multipass AIC/BIC code. Mapper graphs built by the code failed to create edges between nodes generated from non-consecutive intervals even though they shared data points.

\begin{figure}[!htb]
    \centering
    \begin{subfigure}[b]{0.32\textwidth}
        \centering
        \includegraphics[width=.7\textwidth]{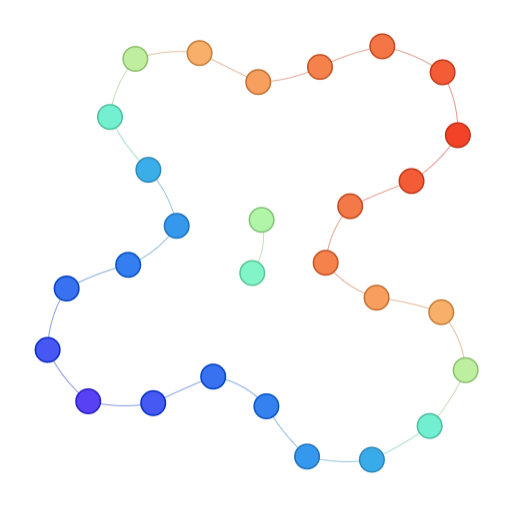}
        \caption{Two Circles Multipass}
        \label{fig:two-circle-multipass}
    \end{subfigure}
    \hfil
    \begin{subfigure}[b]{0.24\textwidth}
        \centering
        \includegraphics[width=.75\textwidth]{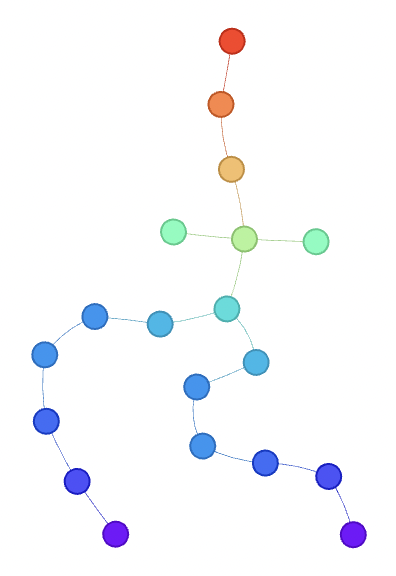}
        \caption{Human Multipass}
        \label{fig:human-multipass}
    \end{subfigure}
    \hfil
    \begin{subfigure}[b]{0.32\textwidth}
        \centering
        \includegraphics[width=.75\textwidth]{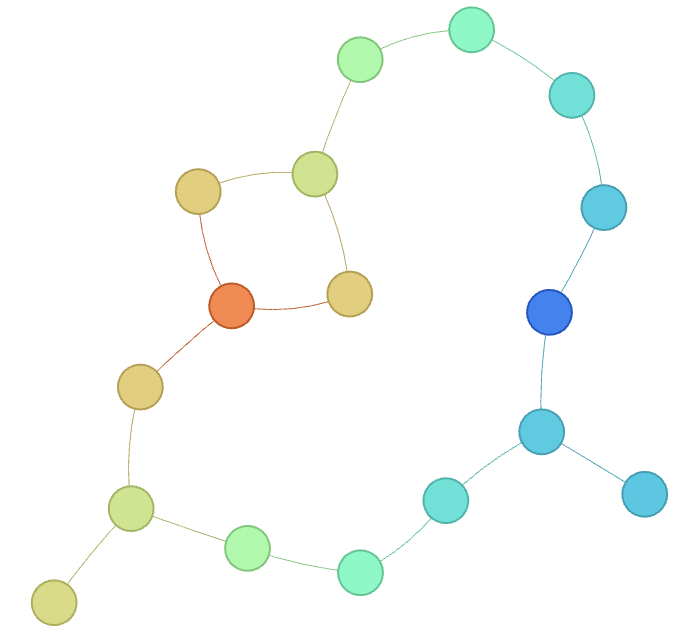}
        \caption{Klein Bottle Multipass}
        \label{fig:klein-multipass}
    \end{subfigure}
    \vskip -10pt
    \caption{Results of Using the Multipass BIC Algorithm on the Synthetic Dataset. \figref{two-circle-multipass} Two Circles: number of initial intervals = 2 and overlap = 0.2. \figref{human-multipass} Human: number of initial intervals = 2 and overlap = 0.2. \figref{klein-multipass} Klein Bottle: number of initial intervals = 4, and overlap = 0.2. The numbers of resulting intervals are 14, 10, and 8, respectively.} 
    \label{fig:multipass-results}
\end{figure}

\begin{figure}[!htb]
    \centering
    \begin{subfigure}[b]{0.32\textwidth}
        \centering
        \includegraphics[width=1\textwidth]{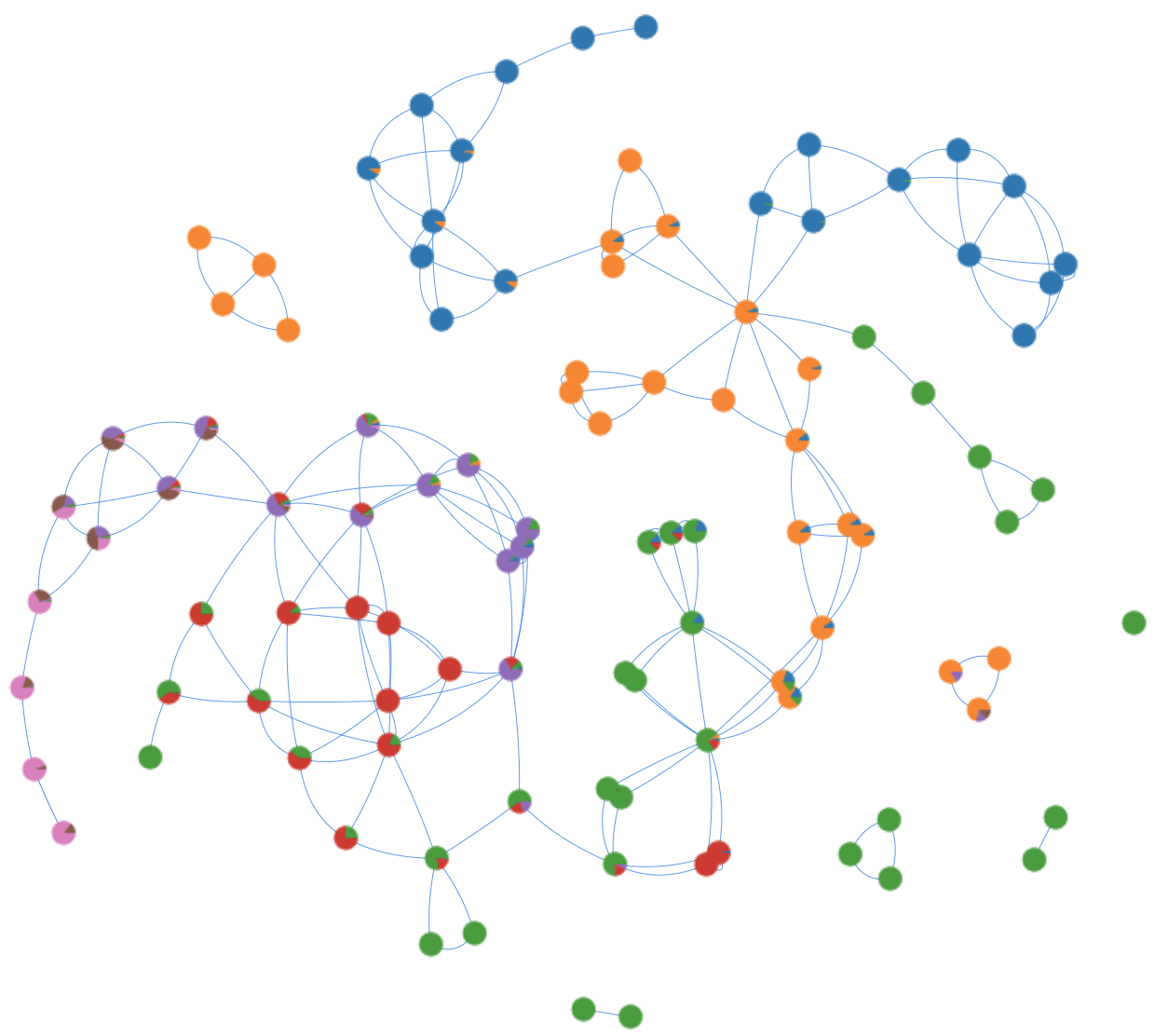}
        \caption{Passiflora Multipass}
        \label{fig:PCA-passiflora-Multipass}
    \end{subfigure}
    \hfil
    \begin{subfigure}[b]{0.32\textwidth}
        \centering
        \includegraphics[width=1\textwidth]{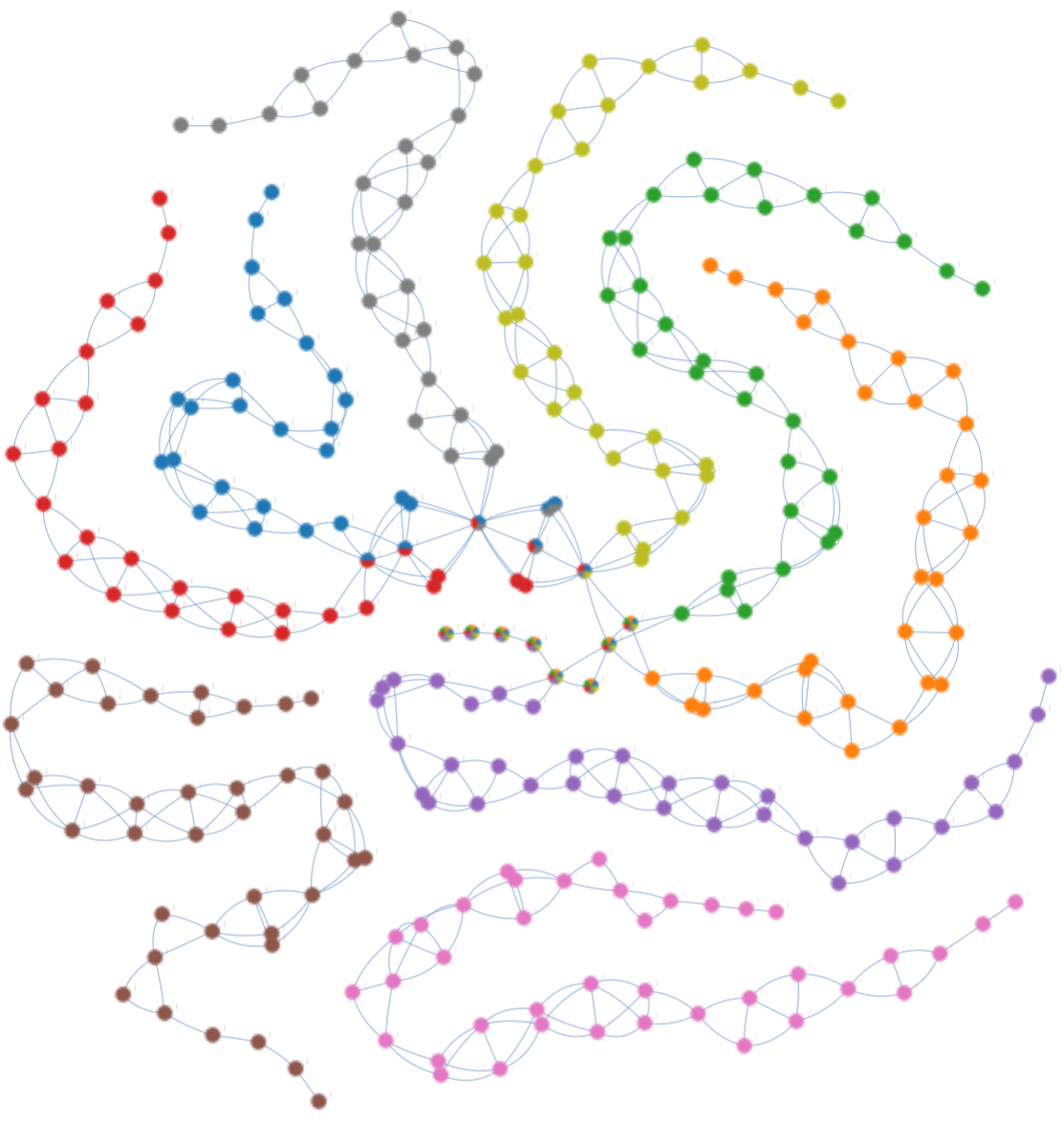}
        \caption{COVID-19 Multipass}
        \label{fig:covid-multipass}
    \end{subfigure}
    \hfil
    \begin{subfigure}[b]{0.32\textwidth}
        \centering
        \includegraphics[width=1\textwidth]{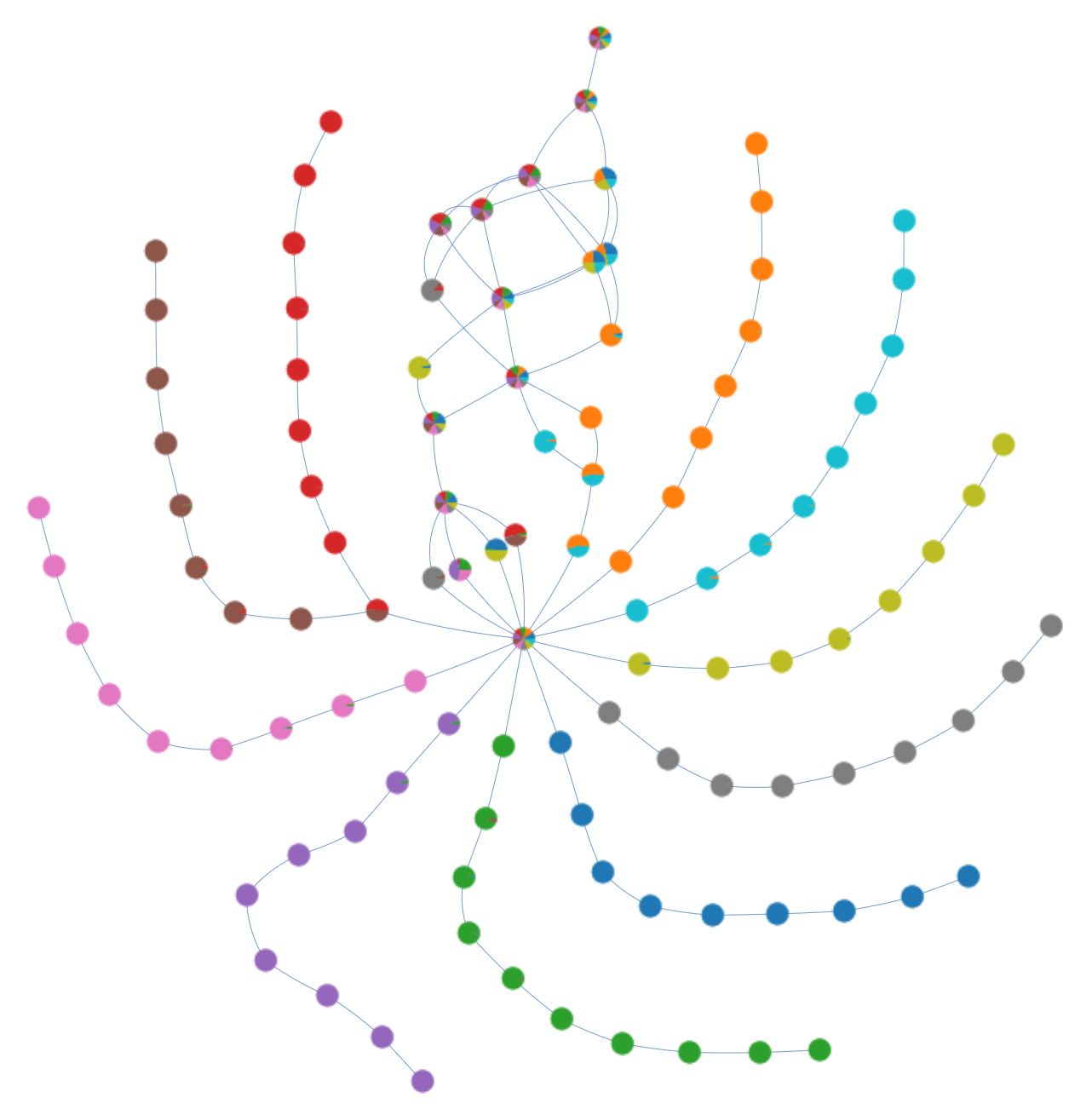}
        \caption{CIFAR-10 Multipass}
        \label{fig:CIFAR-10-Multipass}
    \end{subfigure}
        \vskip -10pt
    \caption{Results of Using the Multipass BIC Algorithm on the Real-World Datasets. \figref{PCA-passiflora-Multipass} Passiflora dataset: number of initial intervals = 3 and overlap = 0.3. \figref{covid-multipass} COVID-19 dataset: the number of initial intervals = 2 and overlap = 0.5.  \figref{CIFAR-10-Multipass} CIFAR-10 dataset: number of initial intervals = 2 and overlap = 0.2. The numbers of resulting intervals are 28, 40, and 22, respectively.}
    \label{fig:multipass-real-data-results}
\end{figure}

\figref{multipass-results} shows that the Multipass BIC algorithm provides a simplified representation of the datasets although the algorithm does not thoroughly capture the essence of each dataset. For the two circles dataset, the inner circle does not appear while the outer circle consists of numerous nodes. For the human dataset, the two arms in the output graph are much shorter than the two legs. Several flares are missing from the Klein bottle output graph. Additionally, we see in the Klein bottle dataset that starting from the numbers of intervals less than the input, $4$, used for generating \figref{klein-multipass} does not lead to splitting intervals. This experiment highlights that choosing the initial number of intervals is a crucial task, which is not required in the G-Mapper algorithm.
\vskip -10pt
    \vskip -20pt
\figref{multipass-real-data-results} represents the Mapper graphs generated by applying the algorithm to the real-world datasets. The Mapper graphs for both the COVID-19 dataset and the Passiflora dataset exhibit similar features to the G-Mapper and reference Mapper graphs, despite having a significantly higher number of nodes. The Mapper graph for the CIFAR-10 dataset separates the 10 different classes, but fails to capture relationships between the classes. In addition, the algorithm splits specific intervals interminably since the data is very high-dimensional, and hence we had to limit the minimum length of intervals.

\subsubsection{F-Mapper}\label{sec:F}
The F-Mapper algorithm finds open intervals based on the fuzzy $c$-means clustering. For the Mapper construction, fuzzy $c$-means is applied to cluster the image $f(X)$ of the data points $X$ under a lens function $f$. For each $p \in f(X)$, the clustering method provides a probability that $p$ belongs to each cluster. To construct intervals from this clustering, the user needs to specify a probability threshold $\tau$. If the probability that $p$ belongs to a specific cluster exceeds $\tau$, then it is declared to be part of the interval containing $p$.

F-Mapper requires the user to specify parameters, including the number of clusters, a threshold value, an exponent value for determining how ``fuzzy" to make the clusters, and an error or convergence parameter. In~\cite{bui2020f}, an exponent value of $2$ and an error of $0.005$ were used for all their examples. The threshold value can be viewed as an overlap parameter. As mentioned before, the number of clusters is difficult to know a priori and is the parameter we are trying to estimate with G-Mapper. 

\begin{figure}[!htb]
    \centering
    \begin{subfigure}[b]{0.32\textwidth}
        \centering
        \includegraphics[width=.7\textwidth]{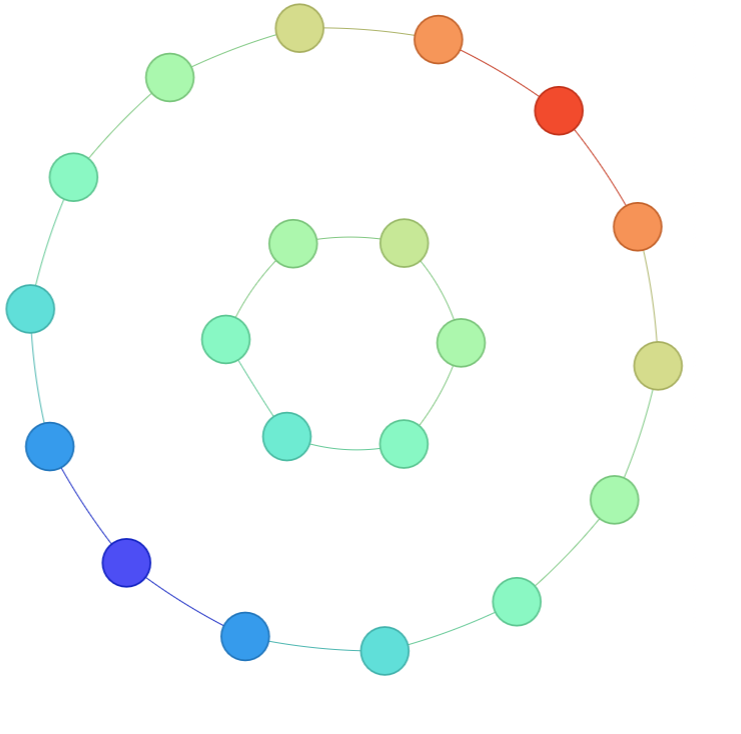}
        \caption{Two Circles F-Mapper}
        \label{fig:two-circle-f}
    \end{subfigure}
    \hfil
    \begin{subfigure}[b]{0.24\textwidth}
        \centering
        \includegraphics[width=.7\textwidth]{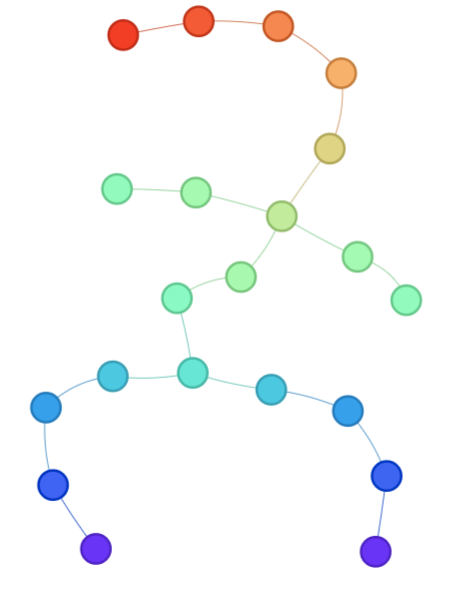}
        \caption{Human F-Mapper}
        \label{fig:human-f}
    \end{subfigure}
    \hfil
    \begin{subfigure}[b]{0.32\textwidth}
        \centering
        \includegraphics[width=.7\textwidth]{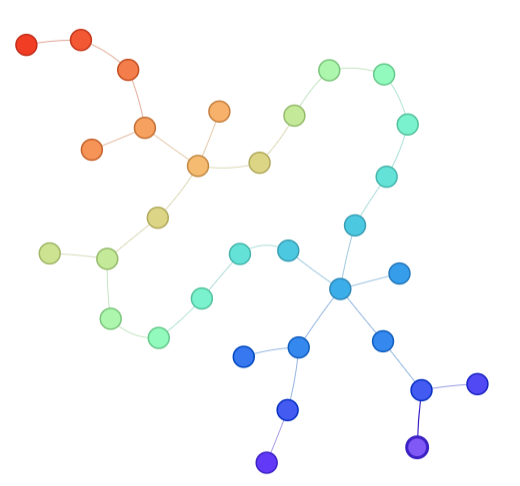}
        \caption{Klein Bottle F-Mapper}
        \label{fig:klein-f}
    \end{subfigure}
        \vskip -10pt
    \caption{Results of Using the F-Mapper Algorithm on the Synthetic Datasets. The numbers of clusters used for \figref{two-circle-f}, \figref{human-f}, and \figref{klein-f} are 8, 12, and 16, respectively.}
    \label{fig:fmapper-results}
\end{figure}

\begin{figure}[!htb]
    \centering
    \begin{subfigure}[b]{0.32\textwidth}
        \centering
        \includegraphics[width=1\textwidth]{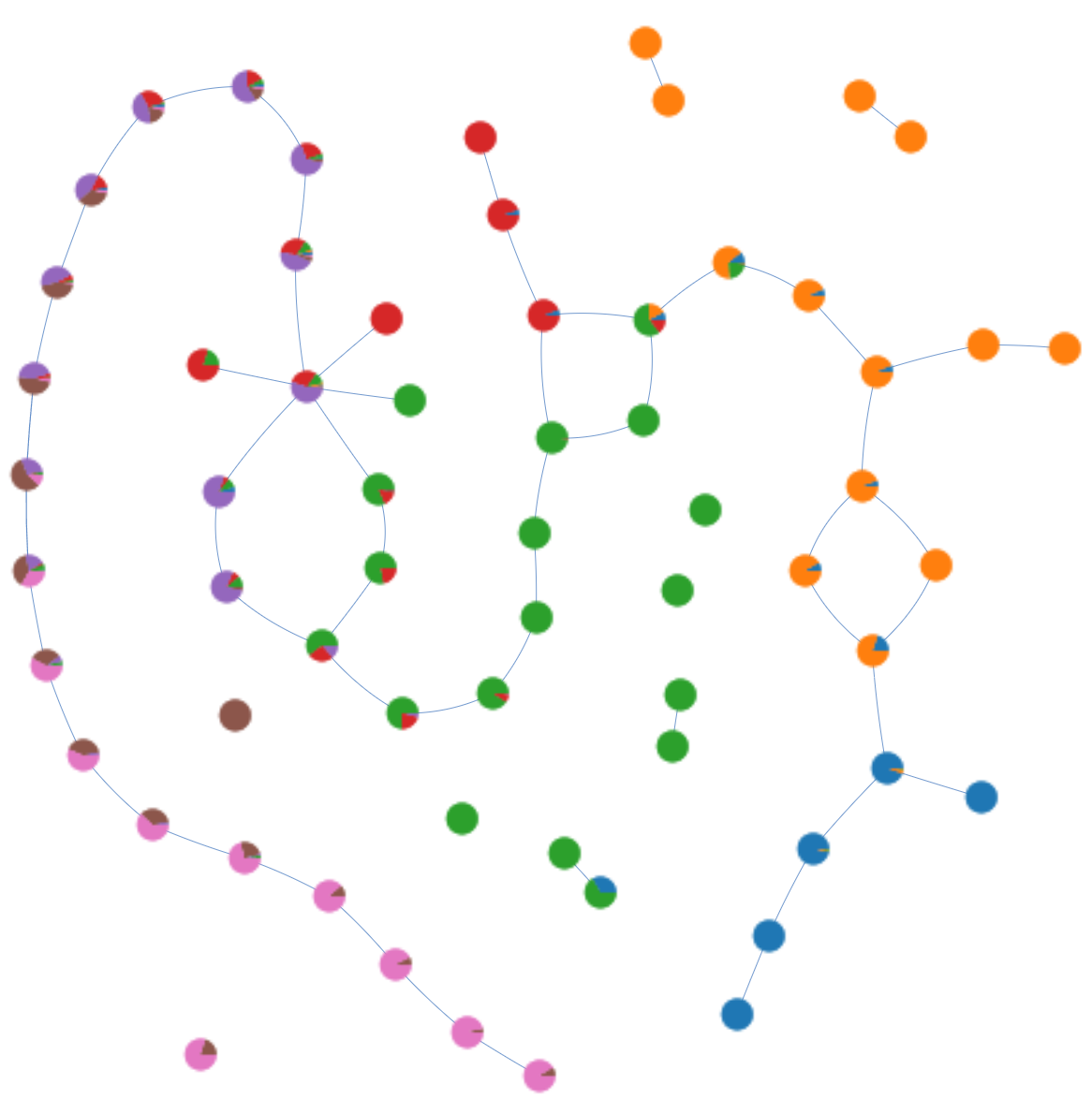}
        \caption{Passiflora F-Mapper}
        \label{fig:Passiflora-fmapper}
    \end{subfigure}
    \hfil
    \begin{subfigure}[b]{0.32\textwidth}
        \centering
        \includegraphics[width=1\textwidth]{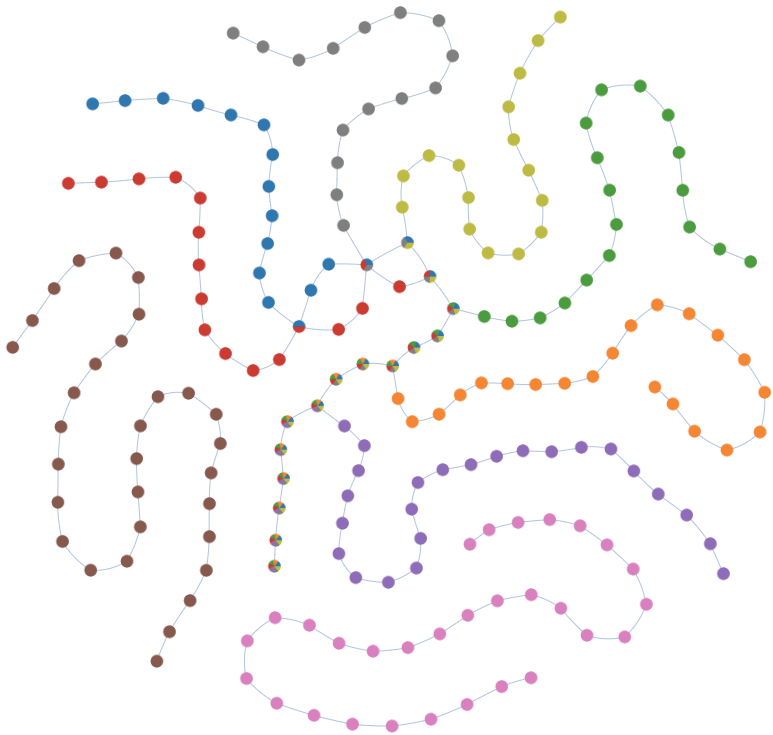}
        \caption{COVID-19 F-Mapper}
        \label{fig:covid-fmapper}
    \end{subfigure}
        \hfil
    \begin{subfigure}[b]{0.32\textwidth}
        \centering
        \includegraphics[width=1\textwidth]{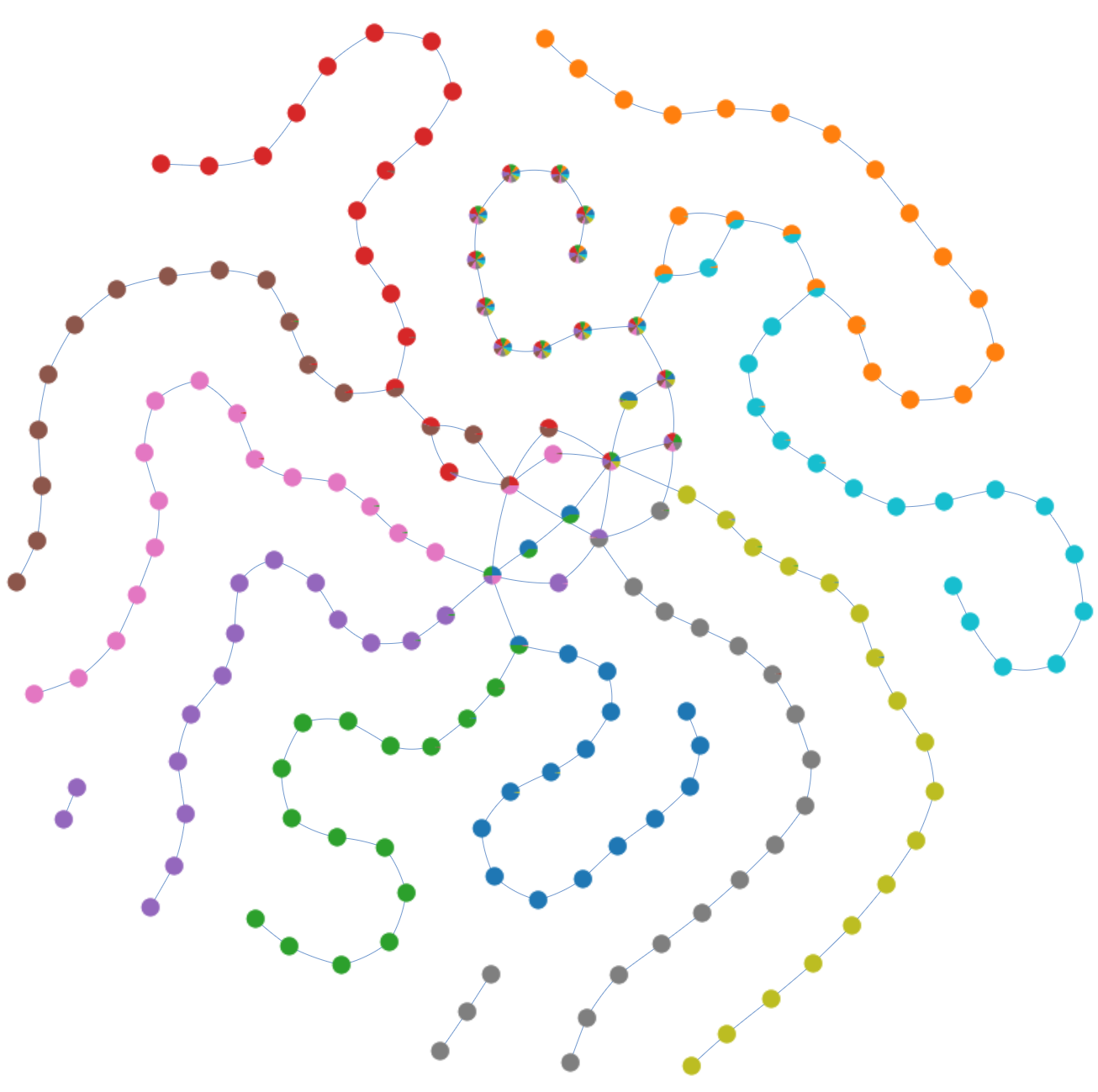}
        \caption{CIFAR-10 F-Mapper}
        \label{fig:CIFAR-10-fmapper}
    \end{subfigure}
        \vskip -10pt
    \caption{Results of Using the F-Mapper Algorithm on the Real-World Datasets. \figref{Passiflora-fmapper} Passiflora dataset: number of intervals = 38. \figref{covid-fmapper} COVID-19 dataset: number of intervals = 31. \figref{CIFAR-10-fmapper} CIFAR-10 dataset: number of intervals = 33.}
    \label{fig:fmapper-realdata-results}
\end{figure}

As an input parameter for the F-Mapper algorithm, we picked the number of intervals obtained from G-Mapper and applied the algorithm to each dataset except for the human dataset. For its implementation, we used the Python package SciKit-Fuzzy 0.4.2 to implement F-Mapper. The results and parameters for F-Mapper are given in \figref{fmapper-results} and \figref{fmapper-realdata-results}. These figures illustrate that F-Mapper with our selective inputs produces Mapper graphs that are almost identical to the reference Mapper Graphs. These examples suggest that G-Mapper could be a useful tool to determine the number of intervals as its input parameter for F-Mapper. 

    \vskip -10pt

\subsubsection{Balanced Cover}\label{sec:balanced}
The balanced cover strategy forms open intervals so that each interval contains the same number of points. To apply the method, the user must specify the number of intervals. When we choose the number of intervals found using G-Mapper as the input for the balanced cover method as done in F-Mapper, it does not always generate the desired Mapper graph because of differences in constructing covers. Hence, we start from the number of intervals obtained by G-Mapper and adjust this number in order to decide the optimal input number of intervals for the balanced cover strategy.

\begin{figure}[!htb]
    \centering
    \begin{subfigure}[b]{0.32\textwidth}
        \centering
        \includegraphics[width=.7\textwidth]{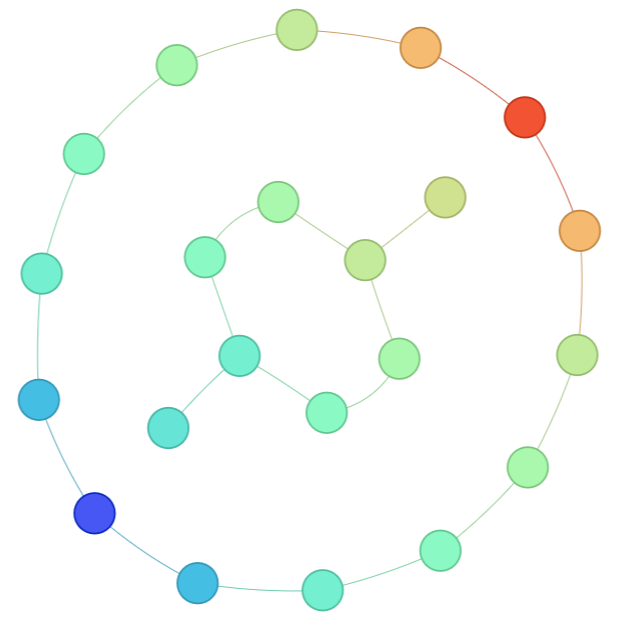}
        \caption{Two Circles Balanced Cover}
        \label{fig:two-circle-balanced}
    \end{subfigure}
    \hfil
    \begin{subfigure}[b]{0.32\textwidth}
        \centering
        \includegraphics[width=.7\textwidth]{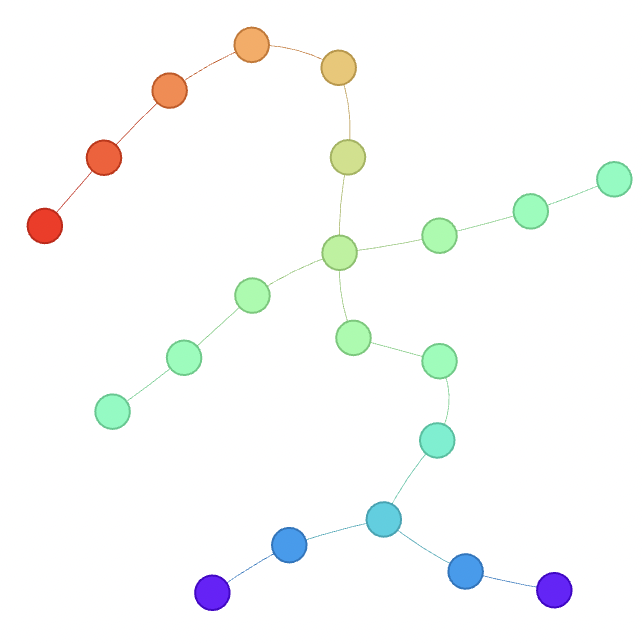}
        \caption{Human Balanced Cover}
        \label{fig:human-balanced}
    \end{subfigure}
    \hfil
    \begin{subfigure}[b]{0.32\textwidth}
        \centering
        \includegraphics[width=.7\textwidth]{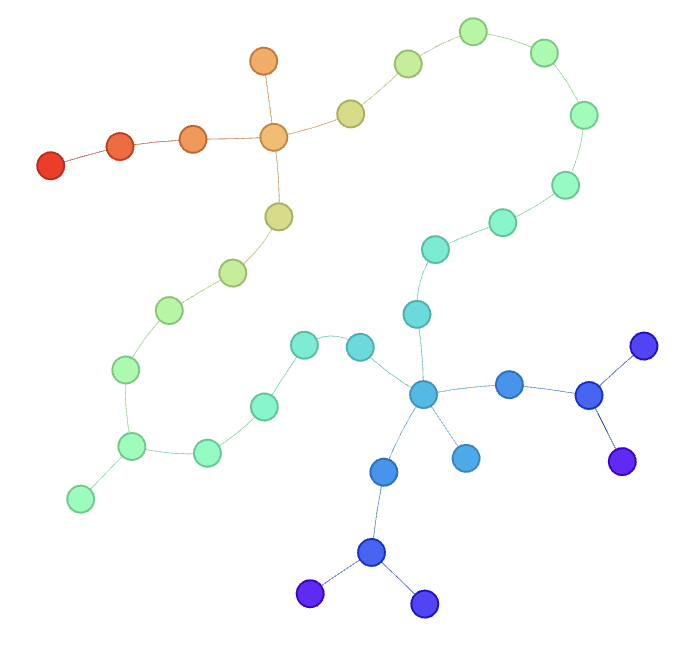}
        \caption{Klein Bottle Balanced Cover}
        \label{fig:klein-balanced}
    \end{subfigure}    \vskip -10pt
    \caption{Results of Using the Balanced Cover Strategy on the Synthetic Datasets. The numbers of intervals used for \figref{two-circle-balanced}, \figref{human-balanced}, and \figref{klein-balanced} are 8, 13, and 17, respectively.}
    \label{fig:balanced-results}
\end{figure}

\begin{figure}[!htb]
    \centering
     \begin{subfigure}[b]{0.32\textwidth}
        \centering
        \includegraphics[width=1\textwidth]{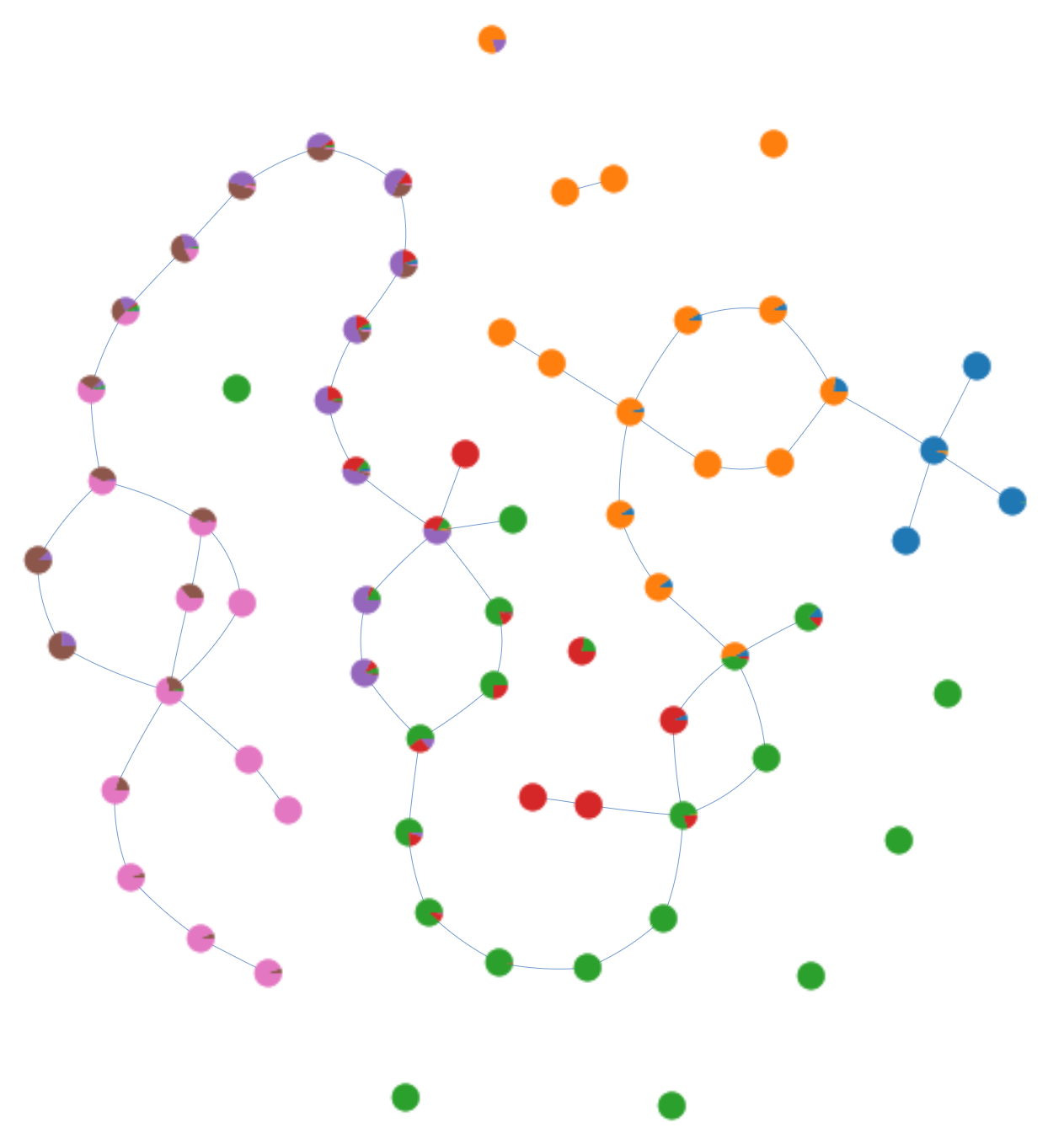}
        \caption{Passiflora Balanced Cover}
        \label{fig:PCA-passiflora-balanced}
    \end{subfigure}
    \hfil
    \begin{subfigure}[b]{0.32\textwidth}
        \centering
        \includegraphics[width=1\textwidth]{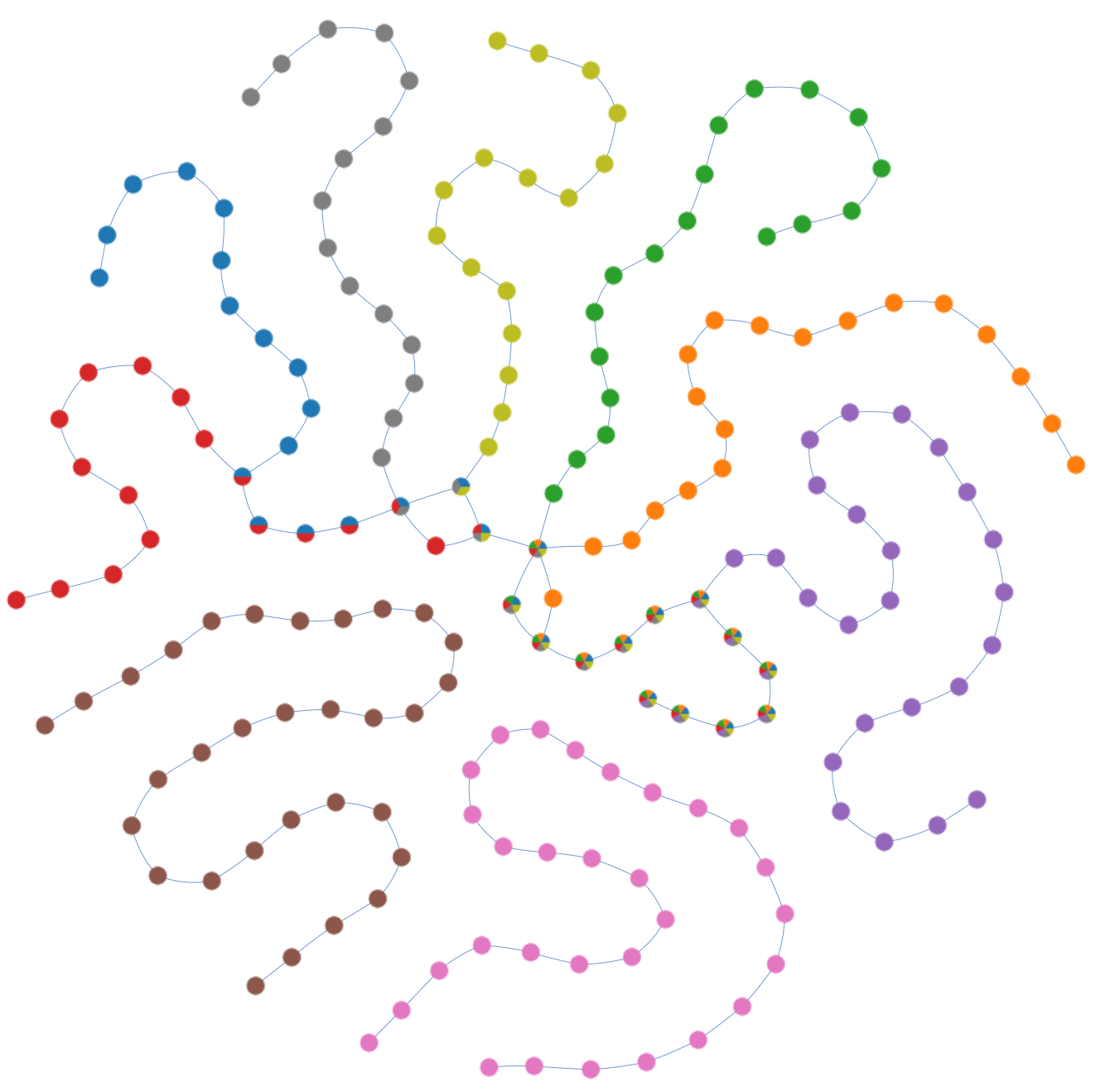}
        \caption{COVID-19 Balanced Cover}
        \label{fig:covid-balanced}
    \end{subfigure}
    \hfil
    \begin{subfigure}[b]{0.32\textwidth}
        \centering
        \includegraphics[width=1\textwidth]{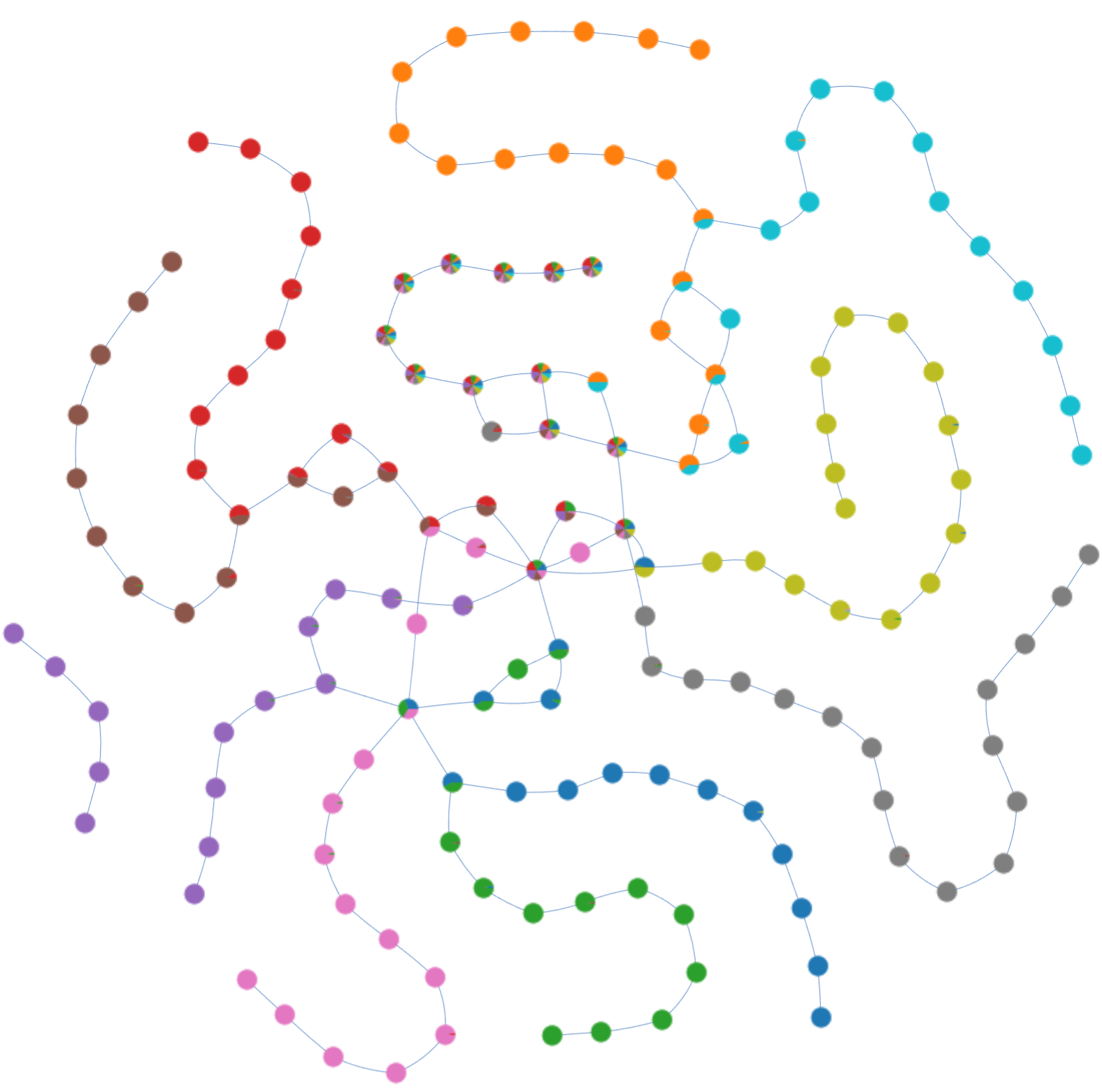}
        \caption{CIFAR-10 Balanced Cover}
        \label{fig:CIFAR-10-balanced}
    \end{subfigure}    \vskip -10pt
    \caption{Results of Using the Balanced Cover Strategy on the Real-World Dataset. \figref{PCA-passiflora-balanced} Passiflora dataset: number of intervals = 38. \figref{covid-balanced} COVID-19 dataset: number of intervals = 31.  \figref{CIFAR-10-balanced} CIFAR-10 dataset: number of intervals = 29.}
    \label{fig:balanced-realdata-results}
\end{figure}

In \figref{balanced-results} and \figref{balanced-realdata-results}, we present Mapper graphs generated by applying the balanced cover strategy to the synthetic and real-world datasets along with the parameters used.
For the synthetic datasets, we chose the number of intervals found using the G-Mapper algorithm except for the human dataset. For that dataset, G-Mapper found 12 intervals but this number results in a disconnected Mapper graph for the Balanced strategy so we opted for 13 intervals instead. For the real-world datasets, we selected the number of intervals derived from G-Mapper except for the CIFAR-10 dataset. For that dataset, G-Mapper found 33 intervals but this number failed to capture the relationship between airplanes (blue points) and birds (green points) for the Balanced strategy so we decided to use 29 intervals instead. \figref{balanced-results} and \figref{balanced-realdata-results} indicate that the balanced cover strategy can generate Mapper graphs similar to both G-Mapper graphs and the reference Mapper graphs although they are not almost identical as in F-Mapper.

    \vskip -10pt

\subsubsection{Runtime Analysis} \label{sec:runtimes}
In \tableref{runtime-results}, we provide the runtimes (in seconds) for generating covers using various Mapper construction algorithms: G-Mapper, Multipass BIC, F-Mapper, and the balanced cover strategy. All the algorithms are executed on a 2.1 GHz 12-core i7 laptop with 32 GB. The times in \tableref{runtime-results} are averaged over five trials and the sizes, and the dimensions of the datasets are also listed in the table.

The G-Mapper algorithm computes a cover faster than the Multipass BIC strategy on all datasets, and is significantly faster for four of the datasets. These four datasets are high-dimensional or have big sizes. The AD statistic is a splitting criterion in G-Mapper and is evaluated on an individual interval. On the other hand, deciding to split an interval for Multipass BIC involves (1) performing a soft clustering on the pre-image of two intervals, if the interval is split for constructing nodes of a Mapper graph, (2) performing a hard clustering on the same set using the previous clustering result, and (3) computing the BIC of the hard clustering. We observe that the first step is the most time-consuming part of the above three steps, dominating the time required for computing the AD statistic in G-Mapper. 

G-Mapper performs faster than F-Mapper for all datasets, and runs significantly faster for the three datasets that are high-dimensional or have big sizes. Hence, utilizing the number of cover intervals generated by G-Mapper as an input parameter for F-Mapper is more effective in regards to runtime than applying F-Mapper multiple times while varying the number of cover intervals. 
\vskip 15pt
\begin{table}[h]
\begin{center}
\begin{tabular}{|c||r|r||r|r||r|}
\hline
\multicolumn{1}{|c|}{\textbf{Dataset}} & \multicolumn{1}{c|}{\textbf{Size}} & \textbf{Dim} & \textbf{G-Mapper} & \textbf{Multipass BIC} & \textbf{F-Mapper}  \\ \hline
Two Circles & 5000 & 2 & 0.196 & 2.186 & 0.150 \\ \hline
Human & 4706 & 3 & 0.142 & 1.232 & 0.220 \\ \hline
Klein Bottle & 15875 & 5 & 0.294 & 7.554 & 12.716  \\ \hline
Passiflora & 3319 & 30 & 0.491 & 58.046 & 2.56  \\ \hline
COVID-19 & 1431 & 7 & 0.207 & 13.260 & 1.439  \\ \hline
CIFAR-10 & 10000  & 512 & 0.488 & 95.851 & 19.238  \\ \hline
\end{tabular}
\caption{Size, dimension, and runtime in seconds of each dataset for each algorithm.}
\label{table:runtime-results}
\end{center}
\end{table}

\subsubsection{Silhouette Values} \label{sec:Silhouettes}
In \tableref{Silhouette-results}, we present the \emph{Silhouette values} \cite{rousseeuw1987silhouettes} for the reference Mapper, G-Mapper, and the previous iterative algorithm (Multipass BIC). The silhouette value is a widely used metric to assess the quality of the clustering, measuring how well each data point is assigned to its cluster. It ranges from $-1$ to $+1$, with higher values indicating that data points are well associated with their own clusters and poorly associated with neighboring clusters. This metric was also employed in \cite{bui2020f} to evaluate the performance of the Mapper algorithm, as Mapper inherently generates a form of soft clustering.

For synthetic datasets, the reference Mapper graphs and the G-Mapper graphs exhibits similar Silhouette values, whereas the Mapper graph generated by Multipass BIC algorithm shows lower values. A similar trend is observed for the Passiflora dataset. For the COVID-19 dataset, the Mapper graph generated by the Multipass BIC algorithm has a significantly lower value, as the graph contains a much larger number of nodes. For the CIFAR-10 dataset, the G-Mapper graph has a higher Silhouette value compared to the other Mapper graphs. Recall that we had to introduce a stopping criterion for the Multipass BIC algorithm due to the high dimensionality of the CIFAR-10 dataset.

Note that since DBSCAN removes outliers in Step 3 of Mapper construction, the resulting datasets differ, which may render the comparisons in this section unfair.
\begin{table}[h]
\begin{center}
\begin{tabular}{|c||r|r|r|}
\hline
\multicolumn{1}{|c|}{\textbf{Dataset}} & \textbf{Reference Mapper} & \textbf{G-Mapper} & \textbf{Multipass BIC} \\ \hline
Two Circles  & $0.236$  & $0.207$  & $0.124$  \\ \hline
Human        & $0.163$  & $0.171$  & $0.086$  \\ \hline
Klein Bottle & $-0.100$ & $-0.092$ & $-0.111$ \\ \hline
Passiflora   & $-0.140$ & $-0.155$ & $-0.171$ \\ \hline
COVID-19     & $0.037$  & $0.006$  & $-0.095$ \\ \hline
CIFAR-10     & $-0.277$ & $-0.253$ & $-0.267$ \\ \hline
\end{tabular}
\caption{The Silhouette value for each dataset and algorithm.}
\label{table:Silhouette-results}
\end{center}
\end{table}

%% file: body/discussion.tex
\section{Discussion}

We have proposed G-Mapper, a novel method for optimizing the cover parameter of a Mapper graph that is motivated by G-means clustering. G-Mapper involves an iterative procedure of splitting cover elements using statistical tests and Gaussian mixture models (GMMs). The Multipass AIC/BIC algorithm is based on $X$-means clustering and iteratively splits intervals according to information criteria. These two methods aim to select an optimal number of intervals. Another Mapper construction algorithm, F-Mapper, relies on fuzzy $c$-means clustering. In contrast to the previous two algorithms, F-Mapper requires choosing the number of intervals in advance. To address this issue, we suggest utilizing the number of intervals generated by G-Mapper as an input for F-Mapper, and show that it is effective in our examples (see \secref{F}).

Our experiments (given in \secref{Synthetic} and \secref{Real-World}) for synthetic and real-world datasets reveal that G-Mapper is able to generate Mapper graphs that illustrate key features of the given data. G-Mapper works well even on non-spherical datasets and high-dimensional datasets as does G-means. We found that the G-Mapper algorithm extracts essential parts that are not uncovered by the Multipass BIC algorithm based on information criteria (see \secref{Multipass}). In addition, the runtime comparison (given in \secref{runtimes}) shows that G-Mapper considerably outperforms Multipass BIC. The running time of G-Mapper hardly depends on the dimension and the size of a given dataset while Multipass BIC is overwhelmingly influenced by these factors. 

The G-Mapper method makes use of a GMM for splitting an interval into two overlapping intervals, which provides an elaborate splitting of intervals. Also, this approach enables G-Mapper to start from the whole target space without choosing the initial number of intervals.  In contrast, regardless of the distribution of a given dataset, the conventional Mapper employs uniform covers, and the Multipass AIC/BIC uniformly splits intervals and requires initialization of a cover. We utilize the means and the variances derived from the GMM for designing two overlapping intervals (refer to Section \ref{sec:G-Mapper_Algorithm}). Considering the weight of each mixture component would give a more specified splitting.

We provide some pointers on optimizing a cover for high-dimensional lens functions $f$, where $f(X) \subseteq \mathbb{R}^n$. To handle the high-dimensional case, the G-Mapper algorithm can be applied separately to each dimension. That is, for $f:X \to \mathbb{R}^n$, we apply G-Mapper to each coordinate function. The resulting cover is then the Cartesian product of the intervals, creating a hypercube. Beyond hypercubes, utilizing high-
dimensional connected sets could yield more desirable covers, but this approach requires a generalization of
the G-Mapper algorithm. When splitting an interval in G-Mapper, only the means and standard deviations from the GMM are utilized. Since this information is insufficient for designing a high-dimensional cover, further exploration is needed to construct the cover, incorporating the covariances and weights from the GMM.

Various clustering algorithms other than G-means, $X$-means, and fuzzy $c$-means may also be suitable to optimize covers in the Mapper construction. Gaussian mixture models with identifying the number of clusters \cite{leroux1992consistent, roeder1997practical, mclachlan2014number} can be applied directly to the image of a lens function instead of our proposed iterative method. The idea behind these techniques is to fit a $k$-component mixture model for varying $k$, and assess the model using a score such as BIC. This approach is more computationally expensive than the proposed algorithm that splits the interval into two intervals since the time complexity of fitting a GMM increases as the number $k$ of components increases. Spectral clustering \cite{von2007tutorial} uses the connectivity of data points instead of compactness like $k$-means, and its fuzzy versions can be found in \cite{roblitz2013fuzzy, wahl2015hierarchical}. Agglomerative clustering \cite{johnson1967hierarchical,jain1988algorithms} is a hierarchical algorithm that recursively merges the closest pairs of clusters, and its fuzzy version was established in \cite{konkol2015fuzzy}. A recent clustering algorithm \cite{cohesionBerenhaut22} introduces a novel technique, partitioned local depth cohesion, interpreted as cohesion, but its soft clustering version has not been developed.

We compared different methods for selecting a cover based on qualitative analysis, runtime, and Silhouette values. We leave making more quantitative comparisons between the Mapper outputs of the different methods as future work. A co-optimal transport metric between hypergraphs is proposed in \cite{chowdhury2023hypergraph}. Since Mapper graphs can be viewed as hypergraphs, the metric is suitable for comparing different Mapper graph outputs from each cover selection method as it was utilized in \cite{rathore2023topobert,zhou2023comparing}.

%% file: article.bbl
\begin{thebibliography}{10}

\bibitem{anderson1952asymptotic}
{\sc T.~W. Anderson and D.~A. Darling}, {\em Asymptotic theory of certain
  ``goodness of fit'' criteria based on stochastic processes}, The annals of
  mathematical statistics,  (1952), pp.~193--212.

\bibitem{cohesionBerenhaut22}
{\sc K.~S. Berenhaut, K.~E. Moore, and R.~L. Melvin}, {\em A social perspective
  on perceived distances reveals deep community structure}, Proceedings of the
  National Academy of Sciences, 119 (2022), p.~e2003634119.

\bibitem{bezdek2013pattern}
{\sc J.~C. Bezdek}, {\em Pattern Recognition with Fuzzy Objective Function
  Algorithms}, Springer Science \& Business Media, 2013.

\bibitem{bishop2006pattern}
{\sc C.~M. Bishop}, {\em Pattern Recognition and Machine Learning}, Springer,
  2006.

\bibitem{bui2020f}
{\sc Q.-T. Bui, B.~Vo, H.-A.~N. Do, N.~Q.~V. Hung, and V.~Snasel}, {\em
  F-{Ma}pper: A fuzzy {Ma}pper clustering algorithm}, Knowledge-Based Systems,
  189 (2020), p.~105107.

\bibitem{carlsson2009topology}
{\sc G.~Carlsson}, {\em Topology and data}, Bulletin of the American
  Mathematical Society, 46 (2009), pp.~255--308.

\bibitem{carriere2018statistical}
{\sc M.~Carri\`{e}re, B.~Michel, and S.~Oudot}, {\em Statistical analysis and
  parameter selection for {M}apper}, The Journal of Machine Learning Research,
  19 (2018), pp.~478--516.

\bibitem{chalapathi2021adaptive}
{\sc N.~Chalapathi, Y.~Zhou, and B.~Wang}, {\em Adaptive covers for {Ma}pper
  graphs using information criteria}, in 2021 IEEE International Conference on
  Big Data (Big Data), IEEE, 2021, pp.~3789--3800.

\bibitem{chen2009benchmark}
{\sc X.~Chen, A.~Golovinskiy, and T.~Funkhouser}, {\em A benchmark for {3D}
  mesh segmentation}, ACM Trans. Graph., 28 (2009).

\bibitem{morphology}
{\sc D.~H. Chitwood and W.~C. Otoni}, {\em {Morphometric analysis of
  {Passiflora} Leaves: the Relationship Between Landmarks of the Vasculature
  and Elliptical {Fourier} Descriptors of the Blade}}, GigaScience, 6 (2017).

\bibitem{chowdhury2023hypergraph}
{\sc S.~Chowdhury, T.~Needham, E.~Semrad, B.~Wang, and Y.~Zhou}, {\em
  Hypergraph co-optimal transport: Metric and categorical properties}, Journal
  of Applied and Computational Topology,  (2023), pp.~1--60.

\bibitem{cohen2009extending}
{\sc D.~Cohen-Steiner, H.~Edelsbrunner, and J.~Harer}, {\em Extending
  persistence using {Poincar{\'e}} and {Lefschetz} duality}, Foundations of
  Computational Mathematics, 9 (2009), pp.~79--103.

\bibitem{dong2020interactive}
{\sc E.~Dong, H.~Du, and L.~Gardner}, {\em An interactive web-based dashboard
  to track {COVID-19} in real time}, The Lancet infectious diseases, 20 (2020),
  pp.~533--534.

\bibitem{duda2012pattern}
{\sc R.~O. Duda, P.~E. Hart, et~al.}, {\em Pattern Classification}, John Wiley
  \& Sons, 2012.

\bibitem{dunn1973fuzzy}
{\sc J.~C. Dunn}, {\em A fuzzy relative of the {ISODATA} process and its use in
  detecting compact well-separated clusters}, Journal of Cybernetics, 3 (1973).

\bibitem{d2017tests}
{\sc R.~B. D’Agostino}, {\em Tests for the normal distribution}, in
  Goodness-of-fit Techniques, Routledge, 2017, pp.~367--420.

\bibitem{ester1996density}
{\sc M.~Ester, H.-P. Kriegel, J.~Sander, X.~Xu, et~al.}, {\em A density-based
  algorithm for discovering clusters in large spatial databases with noise}, in
  kdd, 1996, pp.~226--231.

\bibitem{Fesser2023}
{\sc L.~Fesser, S.~Serrano~de Haro~Iv{\'a}{\~n}ez, K.~Devriendt, M.~Weber, and
  R.~Lambiotte}, {\em Augmentations of {Forman's} {Ricci} curvature and their
  applications in community detections}, arXiv preprint arXiv:2306.06474,
  (2023).

\bibitem{hamerly2003learning}
{\sc G.~Hamerly and C.~Elkan}, {\em Learning the k in k-means}, Advances in
  neural information processing systems, 16 (2003).

\bibitem{hu2004investigation}
{\sc X.~Hu and L.~Xu}, {\em Investigation on several model selection criteria
  for determining the number of cluster}, Neural Information Processing-Letters
  and Reviews, 4 (2004), pp.~1--10.

\bibitem{ikotun2023k}
{\sc A.~M. Ikotun, A.~E. Ezugwu, L.~Abualigah, B.~Abuhaija, and J.~Heming},
  {\em K-means clustering algorithms: A comprehensive review, variants
  analysis, and advances in the era of big data}, Information Sciences, 622
  (2023), pp.~178--210.

\bibitem{jain1988algorithms}
{\sc A.~K. Jain and R.~C. Dubes}, {\em Algorithms for Clustering Data},
  Prentice-Hall, Inc., 1988.

\bibitem{johnson1967hierarchical}
{\sc S.~C. Johnson}, {\em Hierarchical clustering schemes}, Psychometrika, 32
  (1967), pp.~241--254.

\bibitem{konkol2015fuzzy}
{\sc M.~Konkol}, {\em Fuzzy agglomerative clustering}, in Artificial
  Intelligence and Soft Computing: 14th International Conference, ICAISC 2015,
  Zakopane, Poland, June 14-18, 2015, Proceedings, Part I 14, Springer, 2015,
  pp.~207--217.

\bibitem{krizhevsky2009learning}
{\sc A.~Krizhevsky, G.~Hinton, et~al.}, {\em Learning multiple layers of
  features from tiny images}, Technical Report,  (2009).

\bibitem{leroux1992consistent}
{\sc B.~G. Leroux}, {\em Consistent estimation of a mixing distribution}, The
  Annals of Statistics,  (1992), pp.~1350--1360.

\bibitem{li2015identification}
{\sc L.~Li, W.-Y. Cheng, B.~S. Glicksberg, O.~Gottesman, R.~Tamler, R.~Chen,
  E.~P. Bottinger, and J.~T. Dudley}, {\em Identification of type 2 diabetes
  subgroups through topological analysis of patient similarity}, Science
  translational medicine, 7 (2015), pp.~311ra174--311ra174.

\bibitem{clement2014gudhi}
{\sc C.~Maria, J.-D. Boissonnat, M.~Glisse, and M.~Yvinec}, {\em The {Gudhi}
  library: Simplicial complexes and persistent homology}, in Mathematical
  Software -- ICMS 2014, H.~Hong and C.~Yap, eds., Berlin, Heidelberg, 2014,
  Springer Berlin Heidelberg, pp.~167--174.

\bibitem{massey1951kolmogorov}
{\sc F.~J. Massey~Jr}, {\em The {Kolmogorov-Smirnov} test for goodness of fit},
  Journal of the American statistical Association, 46 (1951), pp.~68--78.

\bibitem{mclachlan2014number}
{\sc G.~J. McLachlan and S.~Rathnayake}, {\em On the number of components in a
  gaussian mixture model}, Wiley Interdisciplinary Reviews: Data Mining and
  Knowledge Discovery, 4 (2014), pp.~341--355.

\bibitem{nicolau2011topology}
{\sc M.~Nicolau, A.~J. Levine, and G.~Carlsson}, {\em Topology based data
  analysis identifies a subgroup of breast cancers with a unique mutational
  profile and excellent survival}, Proceedings of the National Academy of
  Sciences, 108 (2011), pp.~7265--7270.

\bibitem{pelleg2000x}
{\sc D.~Pelleg and A.~W. Moore}, {\em X-means: Extending k-means with efficient
  estimation of the number of clusters.}, in Icml, vol.~1, 2000, pp.~727--734.

\bibitem{percival2024topological}
{\sc S.~Percival, J.~G. Onyenedum, D.~H. Chitwood, and A.~Y. Husbands}, {\em
  Topological data analysis reveals core heteroblastic and ontogenetic programs
  embedded in leaves of grapevine (vitaceae) and maracuy{\'a}
  (passifloraceae)}, PLOS Computational Biology, 20 (2024), p.~e1011845.

\bibitem{rathore2021topoact}
{\sc A.~Rathore, N.~Chalapathi, S.~Palande, and B.~Wang}, {\em {TopoAct}:
  {Visually} exploring the shape of activations in deep learning}, in Computer
  Graphics Forum, Wiley Online Library, 2021, pp.~382--397.

\bibitem{rathore2023topobert}
{\sc A.~Rathore, Y.~Zhou, V.~Srikumar, and B.~Wang}, {\em {TopoBERT}: Exploring
  the topology of fine-tuned word representations}, Information Visualization,
  22 (2023), pp.~186--208.

\bibitem{reeb46points}
{\sc G.~Reeb}, {\em Sur les points singuliers d'une forme de {Pfaff}
  completement integrable ou d'une fonction numerique [{On} the singular points
  of a completely integrable {Pfaff} form or of a numerical function]}, Comptes
  Rendus Acad. Sciences Paris, 222 (1946), pp.~847--849.

\bibitem{roblitz2013fuzzy}
{\sc S.~R{\"o}blitz and M.~Weber}, {\em Fuzzy spectral clustering by {PCCA+}:
  Application to {Markov} state models and data classification}, Advances in
  Data Analysis and Classification, 7 (2013), pp.~147--179.

\bibitem{roeder1997practical}
{\sc K.~Roeder and L.~Wasserman}, {\em Practical {B}ayesian density estimation
  using mixtures of normals}, Journal of the American Statistical Association,
  92 (1997), pp.~894--902.

\bibitem{rousseeuw1987silhouettes}
{\sc P.~J. Rousseeuw}, {\em Silhouettes: a graphical aid to the interpretation
  and validation of cluster analysis}, Journal of computational and applied
  mathematics, 20 (1987), pp.~53--65.

\bibitem{shapiro1965analysis}
{\sc S.~S. Shapiro and M.~B. Wilk}, {\em An analysis of variance test for
  normality (complete samples)}, Biometrika, 52 (1965), pp.~591--611.

\bibitem{sinaga2020unsupervised}
{\sc K.~P. Sinaga and M.-S. Yang}, {\em Unsupervised k-means clustering
  algorithm}, IEEE access, 8 (2020), pp.~80716--80727.

\bibitem{singh2007topological}
{\sc G.~Singh, F.~M{\'e}moli, G.~E. Carlsson, et~al.}, {\em Topological methods
  for the analysis of high dimensional data sets and {3D} object recognition.},
  PBG@ Eurographics, 2 (2007).

\bibitem{stephens1974edf}
{\sc M.~A. Stephens}, {\em {EDF} statistics for goodness of fit and some
  comparisons}, Journal of the American statistical Association, 69 (1974),
  pp.~730--737.

\bibitem{tauzin2020giotto}
{\sc G.~Tauzin, U.~Lupo, L.~Tunstall, J.~B. Pérez, M.~Caorsi,
  A.~Medina-Mardones, A.~Dassatti, and K.~Hess}, {\em Giotto-{TDA}: A
  topological data analysis toolkit for machine learning and data exploration},
  Journal of Machine Learning Research, 22 (2020), pp.~1--6.

\bibitem{van2008visualizing}
{\sc L.~Van~der Maaten and G.~Hinton}, {\em Visualizing data using {t-SNE}.},
  Journal of machine learning research, 9 (2008).

\bibitem{van2019kepler}
{\sc H.~J. Van~Veen, N.~Saul, D.~Eargle, and S.~W. Mangham}, {\em Kepler
  {M}apper: A flexible {Python} implementation of the {M}apper algorithm.},
  Journal of Open Source Software, 4 (2019), p.~1315.

\bibitem{vejdemo2020certified}
{\sc M.~Vejdemo-Johansson and A.~Leshchenko}, {\em Certified mapper: Repeated
  testing for acyclicity and obstructions to the nerve lemma}, in Topological
  Data Analysis: The Abel Symposium 2018, Springer, 2020, pp.~491--515.

\bibitem{von2007tutorial}
{\sc U.~Von~Luxburg}, {\em A tutorial on spectral clustering}, Statistics and
  computing, 17 (2007), pp.~395--416.

\bibitem{wahl2015hierarchical}
{\sc S.~Wahl and J.~Sheppard}, {\em Hierarchical fuzzy spectral clustering in
  social networks using spectral characterization}, in The twenty-eighth
  international flairs conference, Citeseer, 2015.

\bibitem{zhou2021mapper}
{\sc Y.~Zhou, N.~Chalapathi, A.~Rathore, Y.~Zhao, and B.~Wang}, {\em Mapper
  interactive: A scalable, extendable, and interactive toolbox for the visual
  exploration of high-dimensional data}, in 2021 IEEE 14th Pacific
  Visualization Symposium (PacificVis), IEEE, 2021, pp.~101--110.

\bibitem{zhou2023comparing}
{\sc Y.~Zhou, H.~Jenne, D.~Brown, M.~Shapiro, B.~Jefferson, C.~Joslyn,
  G.~Henselman-Petrusek, B.~Praggastis, E.~Purvine, and B.~Wang}, {\em
  Comparing {M}apper graphs of artificial neuron activations}, in 2023
  Topological Data Analysis and Visualization (TopoInVis), IEEE, 2023,
  pp.~41--50.

\end{thebibliography}
